\documentclass{article}

\usepackage{PRIMEarxiv}

\usepackage[utf8]{inputenc} 
\usepackage[T1]{fontenc}    

\usepackage{amsmath}        
\usepackage{amssymb}        
\usepackage{amsfonts}       
\usepackage{pifont}
\usepackage{threeparttable}

\usepackage{graphicx}       
\usepackage{float}          
\usepackage{caption}        
\usepackage{subcaption}     

\usepackage{booktabs}       
\usepackage{tabularx}       
\usepackage{multirow}       
\usepackage{adjustbox}      

\usepackage{algorithm}
\usepackage{algpseudocode}

\usepackage{microtype}      
\usepackage{textcomp}       
\usepackage{nicefrac}       
\usepackage{siunitx}        

\usepackage{url}            
\usepackage{hyperref}       
\usepackage{xcolor}         

\usepackage{geometry}       
\usepackage{fancyhdr}       

\usepackage{physics}        

\usepackage{cite}           
\usepackage{lipsum}         
\usepackage{pdfpages}       
\usepackage{orcidlink}      

\usepackage{array}
\newcolumntype{M}[1]{>{\centering\arraybackslash}m{#1}}

\newcommand{\quotes}[1]{``#1''}

\title{QiVC-Net: Quantum-Inspired Variational Convolutional Network, with Application to Biosignal Classification
}

\author{
  Amin Golnari \text{\orcidlink{0000-0002-3841-0588}} \thanks{Corresponding author: Amin Golnari \texttt{<amingolnarii@gmail.com>}} \\
  College of Engineering and Computer Science \\
  VinUniversity \\
  Hanoi, Vietnam \\
  \texttt{amingolnarii@gmail.com} \\
  \And
  Jamileh Yousefi \text{\orcidlink{0000-0003-1901-4819}} \\
  Shannon School of Business \\
  Cape Breton University \\
  Sydney, NS, Canada \\
  \texttt{jamileh\_yousefi@cbu.ca} \\
  \And
  Reza Moheimani \text{\orcidlink{0009-0002-3560-5731}} \\
  Faculty of Computer Science \\
  Chemnitz University of Technology \\
  Chemnitz, Germany \\
  \texttt{rezamoheiman@gmail.com} \\
  \And
  Saeid Sanei \text{\orcidlink{0000-0002-1446-5744}} \\
  College of Engineering \& Computer Science and College of Health Science, VinUniversity, Vietnam \\
  Institute of Psychiatry, Psychology \& Neuroscience, King’s College London, UK \\
  \texttt{saeid.s@vinuni.edu.vn} \\
}

\begin{document}
\maketitle

\begin{abstract}
In this paper, a learning framework is introduced which incorporates principles of probabilistic inference, variational optimization, and geometry-preserving operations inspired by quantum transformations. The central innovation of this quantum-inspired variational convolution (QiVC) lies in its quantum-inspired rotated ensemble (QiRE) mechanism. QiRE performs differentiable low-dimensional subspace rotations of convolutional weights. By drawing a mathematical analogy from unitary evolution, this approach enables structured uncertainty modeling that respects the intrinsic geometry of the parameter space. To demonstrate its practical potential, the concept is instantiated in a QiVC-based convolutional network (QiVC-Net) and evaluated in the context of biosignal classification, focusing on phonocardiogram (PCG) recordings. The proposed QiVC-Net integrates an architecture in which the QiVC layer does not introduce additional parameters, instead performing an ensemble rotation of the convolutional weights through a structured mechanism ensuring robustness without added highly computational burden. Experiments on two benchmark datasets, PhysioNet CinC 2016 and PhysioNet CirCor DigiScope 2022, show that QiVC-Net achieves state-of-the-art performance, reaching accuracies of $97.84\%$ and $97.89\%$, respectively. These findings highlight the versatility of the QiVC framework and its promise for advancing uncertainty-aware modeling in real-world biomedical signal analysis. The implementation of the QiVConv layer is available in \href{https://github.com/amingolnari/Demo-QiVC-Net}{GitHub} for public use.

\end{abstract}


\section{Introduction}

Probabilistic deep learning has emerged as a critical direction for advancing model reliability, interpretability, and robustness in real-world decision-making settings. Unlike deterministic neural networks that produce single-point predictions, probabilistic models explicitly account for uncertainty, allowing them to represent ambiguity in both parameters and observations \cite{gal2016dropout, kendall2017uncertainties}. However, many existing Bayesian deep learning frameworks approximate uncertainty through simplistic isotropic noise models or mean-field assumptions \cite{abdar2021review}, which fail to capture the structured and geometry-aware variability present in high-dimensional parameter spaces. As a result, uncertainty estimates often become poorly calibrated and unstable under distributional shifts, limiting their usefulness in safety-critical applications \cite{krishnan2020improving, taori2020measuring}.

To address these limitations, the quantum-inspired variational convolution (QiVC) framework is introduced, incorporating a geometric and variational treatment of uncertainty directly within convolutional layers. The central innovation of this framework is the quantum-inspired rotated ensemble (QiRE) mechanism, which applies differentiable, low-dimensional subspace rotations to convolutional kernels. This operation is inspired by unitary evolution in quantum systems and preserves the intrinsic geometry and norm of the weight space \cite{sakurai2020modern}. By enforcing coherent structure in stochastic perturbations, the QiVC layer enables richer uncertainty representation while avoiding the instability and expressiveness limitations of isotropic or unstructured Gaussian perturbations. Importantly, QiVC does not introduce additional learnable parameters; instead, it structurally re-parameterizes the convolutional filters to generate expressive, probabilistically grounded ensembles with minimal computational overhead.

Building on prior developments in quantum-inspired neural architectures \cite{beer2020training, li2024quantum} and functional variational inference \cite{sun2019functional}, QiVC provides a principled approach for embedding uncertainty directly into the model’s functional representation. By unifying variational learning with geometry-preserving transformations, it bridges Bayesian inference with structured ensemble modeling, yielding stable and interpretable feature representations suitable for decision-critical domains.

In this work, the QiVC framework is instantiated within a convolutional network, referred to as QiVC-Net, and its effectiveness is evaluated in the context of phonocardiogram (PCG) classification. PCG signal analysis plays a vital role in computer-aided auscultation and early detection of cardiovascular diseases. Heart sound recordings provide clinically significant diagnostic cues, yet their interpretation is complicated by low signal-to-noise ratios, high inter-patient variability, and environmental artifacts typical of real-world settings \cite{sanei2020body}. Recent advances in deep learning have improved PCG classification performance through CNNs, RNNs, transformers, and hybrid models \cite{zhao2024deep, ameen2024advances}. Nevertheless, persistent challenges remain regarding robustness, interpretability, dataset imbalance, and the lack of reliable uncertainty quantification \cite{dwivedi2018algorithms, azam2022cardiac, fernando2021deep}.

To enhance temporal coherence and robustness to physiological variability, QiVC-Net further incorporates a reversal fusion residual (RFR) block that models bidirectional temporal dependencies in PCG recordings. This combination of geometry-aware probabilistic convolution and temporal symmetry makes QiVC-Net well-suited for biosignal analysis under noisy and heterogeneous conditions.

The main contributions of this work are summarized as follows:

\begin{itemize}
    \item \textbf{QiVConv layer:} A probabilistic convolutional layer introducing norm-preserving subspace rotations that enhance uncertainty calibration while maintaining expressive kernel structure.
    \item \textbf{QiRE sampling:} A structured stochastic perturbation mechanism that performs differentiable unitary-inspired rotations of convolutional weights, enabling Bayesian-like inference without complex-valued arithmetic.
    \item \textbf{Uncertainty-aware dual-path architecture:} Integration of QiVConv with a RFR block to capture bidirectional temporal dynamics in PCG signals without adding extra learnable parameters.
    \item \textbf{Clinically aligned design:} A computationally efficient and interpretable framework unifying uncertainty quantification, robustness to signal variation, and temporal coherence for reliable PCG-based decision support.
\end{itemize}

\section{Literature Review}

\subsection{Architectural and Probabilistic Modeling Advances}

The focus of recent research has been to improve the robustness to noise and recording variability, strengthening temporal feature modeling, and developing more reliable uncertainty estimation in PCG classification. Barnawi et al. \cite{barnawi2023simple} introduced a minimal preprocessing strategy combined with CNN fine-tuning to improve efficiency. Riaz et al. \cite{riaz2021novel} developed a portable embedded system leveraging hand-crafted spectral and temporal features. Bidirectional recurrent architectures \cite{schuster1997bidirectional} have been shown to effectively capture the inherent temporal symmetry of PCG signals \cite{latif2018phonocardiographic}, while hybrid CNN–BiLSTM ensembles improve robustness under noisy conditions \cite{kalatehjari2025advanced}. Phoemsuk and Abolghasemi \cite{phoemsuk2025enhanced} further highlighted the importance of signal-quality-aware preprocessing within lightweight one-dimensional CNNs for reliable cardiovascular disease detection.

On the probabilistic side, dropout-based inference \cite{gal2016dropout} and Bayes by Backprop \cite{blundell2015weight} introduced scalable uncertainty modeling, and later approaches such as multiplicative normalizing flows \cite{louizos2017multiplicative} improved posterior expressiveness. However, these frameworks typically overlook the geometric dependencies among kernel parameters, where naive noise injection may violate intrinsic correlations and distort the weight space geometry. Consequently, existing Bayesian formulations often misrepresent epistemic uncertainty in structured signals like PCG.

\subsection{Phonocardiogram Classification}

Over the past decade, PCG classification has progressed from model-driven to data-driven feature engineering toward hybrid and fully deep learning frameworks, with increasing attention to label-efficient, resource-constrained, and murmur-aware designs. Tuncer et al. \cite{tuncer2021application} introduced a graph-theoretic framework combining Petersen graph patterns (PGP), tent energy pooling (TEP), and iterative neighborhood component analysis (INCA) to extract compact yet discriminative features from PCG recordings. Chen et al. \cite{chen2016s1} applied a three-layer deep neural network to mel-frequency cepstral coefficient (MFCC) super-vectors to identify S1 and S2 heart-sound events directly from acoustic data, demonstrating the feasibility of purely audio-based deep models for event-aware classification without explicit temporal priors.

As research advanced, two-dimensional spectrogram representations became a preferred input format for discriminative learning. Karhade et al. \cite{karhade2022time} transformed PCG signals into polynomial-chirplet spectrograms and classified them using a custom CNN. Khan et al. \cite{khan2022cardi} developed Cardi-Net, a deep residual CNN designed for high-resolution power spectrograms with extensive augmentation to enhance robustness. Joshi et al. \cite{joshi2022ai} integrated one-dimensional and two-dimensional CNNs into a handheld, three-dimensional printed stethoscope, enabling on-device PCG analysis with minimal latency.

Although spectrogram-based models offer strong representational capacity, they often require large annotated datasets. Transfer learning has been widely adopted to address this challenge. Hettiarachchi et al. \cite{hettiarachchi2021novel} proposed a dual-input architecture that fuses separately pre-trained CNN models using PCG and electrocardiogram (ECG), enabling effective classification from limited synchronized recordings. Almanifi et al. \cite{almanifi2022heartbeat} demonstrated the adaptation of image-domain pre-trained models to PCG spectrograms for transfer learning. Nehary and Rajan \cite{nehary2024phonocardiogram} explored positive-unlabelled learning for situations with incomplete annotations, using CNN ensembles with self-attention to separate known normal cycles from unlabeled abnormal ones.

To better detect murmurs in raw recordings, several studies have inserted explicit signal separation stages before classification. The foundational principle of isolating crucial cyclic components was demonstrated by Sanei et al. \cite{sanei2011new}, who proposed a singular spectrum analysis-based adaptive line enhancer (SSA-ALE) to robustly separate quasi-periodic signals, such as heart murmurs, from noise by exploiting the full eigen-spectrum of the SSA embedding matrix. Building on this, Qi and Sanei \cite{qi2024murmur} later developed a two-stage murmur isolation method combining constrained SSA with wavelet refinement, guided by acoustic statistics such as zero-crossing rate and kurtosis. This approach enhances the downstream classification and can reduce computational load. In parallel, Ma et al. \cite{ma2024edge} applied self-supervised pretraining on large unlabeled PCG corpora to create efficient one-dimensional CNN models suitable for mobile devices. Arslan \cite{arslan2022automated} combined traditional time-frequency features with CNN-extracted representations through recursive feature elimination and ensemble-based decision making, yielding architectures appropriate for embedded systems.

The methodological diversity has also extended to broader cardiac conditions. Li et al. \cite{li2020fusion} integrated multi-domain hand-crafted features with CNN-derived MFCC embeddings in a hybrid fusion strategy for coronary artery disease detection. Iqtidar et al. \cite{iqtidar2021phonocardiogram} used adaptive local ternary pattern descriptors concatenated with MFCCs after empirical mode decomposition denoising, applying these representations in a machine learning classifier. In the paediatric domain, Oliveira et al. \cite{oliveira2021circor} released the CirCor dataset, offering multi-location recordings with rich murmur annotations and auxiliary descriptors such as timing, shape, and pitch, supporting the development of algorithms robust to diverse recording conditions. These advances collectively reflect a shift from isolated algorithmic benchmarks toward integrated, clinically informed systems capable of supporting real-world cardiac screening.

\subsection{Quantum-Inspired Deep Learning}

Quantum-inspired deep learning (QiDL) is an emerging paradigm that embeds quantum-mechanical principles into classical neural architectures and optimizers. These principles include superposition, entanglement, and unitary evolution. Distinct from full quantum machine learning, which requires quantum hardware, QiDL operates on conventional CPUs and GPUs by simulating quantum phenomena using complex-valued weights, density-matrix representations, and Hilbert-space feature mappings \cite{huynh2023quantum}. Key abstractions in QiDL include superposition analogues, where input vectors are mapped to complex amplitudes. This allows a single neuron to encode an exponential number of states. Entanglement proxies are implemented using product-based layers or graph-coupled gates. These components capture high-order correlations without relying on physical qubits. Finally, a density-matrix formalism replaces conventional activations with positive-semidefinite operators, which naturally quantifies epistemic uncertainty and aligns QiDL with Bayesian inference. These constructions yield richer representational capacity and improved generalization in data-scarce, high-dimensional regimes such as omics, imaging, and longitudinal clinical records.

Optimization techniques span dequantized linear-algebra speedups (Tang-style algorithms), Ising-type energy minimizers \cite{li2021quantum}, and quantum-walk kernels for graph similarity \cite{bai2019quantum}. QiDL’s probabilistic structure dovetails naturally with Bayesian reasoning. Bayesian hyper-parameter tuning deploys quantum-inspired genetic algorithms (QGA) to prune electroencephalogram (EEG) features before feeding a quantum support vector machine (SVM), yielding $57\%$ cross-subject motor-imagery accuracy with $p<0.03$ significance while using fewer samples than classical SVM \cite{lalawat2025advancing}. Density-matrix layers output quantum-entropic uncertainties that are propagated through Bayesian neural networks (BNN), improving calibration in low-data drug-efficacy trials \cite{li2021quantum}. Energy-based priors derived from Ising Hamiltonians act as implicit regularizers, shrinking the effective parameter space and mitigating over-fitting in rare-disease genomics. These formulations enable enhanced modeling of uncertainty, relational structure, and non-local dependencies, offering a principled framework for inference under noise and partial observability.

The use of Hilbert space embeddings allows for high-dimensional feature representations that capture subtle patterns in the data, while complex-valued activations facilitate phase-sensitive interference effects analogous to quantum coherence. However, unlike complex-valued architectures that incur significant memory overhead, QiVC-Net operates efficiently within a real-valued vector space. By utilizing real-valued QR decompositions and orthogonal Haar rotations, the framework successfully constructs a computationally lightweight, real-space analog of norm-preserving quantum evolution to enhance sensitivity to semantic and structural nuances in biomedical signals. Furthermore, quantum-inspired similarity measures based on fidelity between probabilistic state representations have demonstrated superior performance in capturing global structural relationships \cite{bai2019quantum}. This advantage is especially evident in sequential and graph-structured data. Together, these synergies enhance robustness, interpretability, and sample efficiency, three critical attributes for clinical deployment.

\begin{table}[H]
\centering
\caption{Impact of quantum-inspired models in biomedical applications.}
\label{table:qi_biomedicine}

\begin{adjustbox}{max width=\textwidth}
\begin{tabular}{l M{3.5cm} M{4.5cm} M{4.5cm}}
\toprule
\textbf{Domain} & \textbf{Model} & \textbf{Task} & \textbf{Key Result} \\
\midrule

Psychiatry & QGA-QSVM \cite{lalawat2025advancing} & Depression screening from EEG & Statistically superior with 20\% less training data \\[2mm]

Endocrinology & QuCNet \cite{swathi2024qucnet} & Thyroid-nodule malignancy & 97.6\% accuracy on limited ultrasound data \\[2mm]

Cardiology & QiEA-SMGLR \cite{siddiqui2024towards} & Denoising phonocardiograms & 100\% max SNR improvement, $\rho\approx0.99$ clinical correlation \\[2mm]

Clinical NLP & QPFE-ERNIE \cite{shi2024pretrained} & Sentiment \& word-sense disambiguation in EHR notes & $+8.7$ F1 over prior quantum models \\[2mm]

Neuro-imaging & IDLQET-BTEDC \cite{alamri2025innovative} & Brain-tumor edge detection \& classification & 98\% accuracy, Dice=89\%, superior to DeepMedic \\
\bottomrule
\end{tabular}
\end{adjustbox}
\end{table}

As indicated in Table \ref{table:qi_biomedicine}, these studies collectively show consistent improvements in accuracy, convergence speed, and noise resilience across modalities where the data are scarce, high-dimensional, or privacy-sensitive. Nevertheless, several challenges remain. These challenges and emerging directions characterize current research in quantum-inspired deep learning:

\begin{itemize}
    \item \textbf{Computational overhead:} Simulating entanglement on classical hardware incurs computational cost. Tensor-network approximations can partially mitigate this bottleneck but do not eliminate it entirely \cite{huynh2023quantum}.
    
    \item \textbf{Interpretability:} Complex-valued weights and entanglement proxies pose challenges for human interpretability. Recent studies explore quantum-circuit visualizations and entanglement-entropy heat maps as tools for better model explainability \cite{shi2024pretrained}.
    
    \item \textbf{Hardware limitations:} Physical quantum accelerators remain prone to noise and decoherence. Hybrid pipelines have been proposed, in which only the optimization layer is offloaded to near-term quantum processing units (QPUs) while inference remains classical \cite{ren2025toward}.
    
    \item \textbf{Future directions:} Promising research avenues include pretrained quantum-inspired transformers for multi-modal electronic health record (EHR) fusion, quantum federated learning (QFL) using blind quantum computation and quantum key distribution for privacy-preserving training across hospitals \cite{ren2025toward}, and Bayesian–quantum co-design approaches where hyperpriors are optimized via variational quantum eigensolvers and the model uncertainty is back-propagated through quantum-aware layers.
\end{itemize}

As QiDL matures, the confluence of Bayesian inference, quantum metaphors, and classical deep learning is poised to deliver robust, sample-efficient solutions for precision medicine and population health analytics. Table \ref{tab:qi_summary} summarizes the key quantum-inspired learning paradigms, highlighting their core computational mechanisms, empirical gains across diverse tasks, and the practical bottlenecks reported in the literature. This integrated overview illustrates how different families of models range from tensor-network approaches to quantum-inspired CNNs, DBNs, GNNs, and pre-trained NLP architectures. These models balance theoretical advantages with implementation constraints. The summary provides a foundation for informed selection and further development of QiDL architectures.

\begin{table}[H]
\centering
\caption{Summary of quantum-inspired models: core mechanisms, empirical gains, and practical bottlenecks.}
\label{tab:qi_summary}

\begin{adjustbox}{max width=\textwidth}
\begin{tabular}{l M{2.7cm} M{3.5cm} M{4.5cm} M{5.5cm}}
\toprule
\textbf{Family} & \textbf{Model} & \textbf{Core Mechanism} & \textbf{Reported Advantage} & \textbf{Acknowledged Disadvantage} \\
\midrule

\multirow{1}{*}{Tensor-Network Models} & \multirow{1}{*}{MPS/TTN \cite{huynh2023quantum}} 
& Low-rank tensor contractions emulating exponentially large Hilbert spaces
& Polynomial runtime for tasks classically NP-hard; exponential parameter compression
& Cubic-to-quartic scaling in bond dimension $D$; collapses on high-res images or large vocabularies; slower than CNN/Transformer at realistic sizes
\\[2mm]

\multirow{1}{*}{Quantum-Inspired CNNs} & \multirow{1}{*}{QuCNet \cite{swathi2024qucnet}}
& Random quantum-circuit filters applied to $2\times2$ quanvolutional patches
& 97.6\% thyroid-nodule accuracy with 5–30\% fewer parameters vs.\ DenseNet121
& 8.15 h GPU training; performance degrades with $>$2 circuit repetitions; exponential cost in kernel size forces tiny $2\times2$ kernels
\\[2mm]

\multirow{1}{*}{Quantum-Inspired GNNs} & \multirow{1}{*}{QiEA-SMGLR \cite{siddiqui2024towards}}
& Sorted-mutation evolutionary search over FIR-filter coefficients
& 20–100\% SNR gain in PCG denoising; $\rho\approx0.99$ clinical fidelity
& Per-generation filter evaluation → heavy CPU load; tiny 6-clip study; fixed 4-tap filter cannot handle richer spectra without exploding search space
\\[2mm]

\multirow{1}{*}{Pre-trained NLP} & \multirow{1}{*}{QPFE-ERNIE \cite{shi2024pretrained}}
& ERNIE embeddings fused with complex-valued GRU “evolution circuits”
& $+8.7$ F1 on sentiment tasks; interpretable quantum phase parameters
& Training cost doubles (two forward passes per batch); fragile hand-tuned hyper-parameters; extra pre-training stage required
\\[2mm]

\multirow{1}{*}{Quantum-Inspired DBNs} & \multirow{1}{*}{QISOA-DBN \cite{banumathy2024quantum}}
& Seagull optimiser with qubit-rotation gates and superposition states
& 98.6\% CVD accuracy; 3$\times$ faster convergence than grid-search baselines
& Resource-hungry five-layer DBN prone to over-fitting; doubtful transferability to non-cardiac domains; high energy footprint
\\
\bottomrule
\end{tabular}
\end{adjustbox}
\end{table}

Building upon the foundational principles of quantum-inspired deep learning, this study extends recent probabilistic and architectural advances by introducing a framework that embeds structured stochasticity into the learning process. Our approach draws inspiration from unitary transformations and density-based representations to ensure consistency between uncertainty modeling and temporal feature extraction. This integration, detailed in the following section, aims to address persistent challenges in biomedical time-series analysis such as data scarcity, noise sensitivity, and interpretable uncertainty estimation.

\section{Datasets}

This study employs two publicly available PCG datasets, the PhysioNet/computing in cardiology challenge 2016 dataset (CinC) \cite{liu2016classification} and the CirCor DigiScope cataset 2022 (CirCor) \cite{oliveira2022circor}. Both datasets provide real-world heart sound recordings with clinical annotations, but they differ notably in demographic composition, acquisition protocols, labeling schemes, and structural organization. These diverse benchmarks were specifically chosen to rigorously evaluate uncertainty calibration and model robustness across varying signal qualities and demographic distributions, ranging from controlled adult recordings to noisy pediatric screenings. These differences present complementary challenges for developing and evaluating robust and portable cardiac sound analysis methods.

\subsection{PhysioNet/CinC Challenge 2016 Dataset (CinC)}

The CinC dataset, introduced as part of the PhysioNet Computing in Cardiology Challenge 2016, aggregates PCG recordings from multiple international sources. The recordings were collected in diverse settings, ranging from controlled clinical environments to unstructured non-clinical contexts, using a variety of digital stethoscopes and mobile devices. Participants' ages span from $18$ to $85$ years, with a near-balanced gender distribution. This heterogeneity introduces real-world variability in recording quality and background interference, making the dataset particularly valuable for evaluating model robustness under practical conditions.

The dataset is organized into six labeled subsets, \textit{training-a} through \textit{training-f}, contributed by different institutions. Only \textit{training-a} contains actual PCG recordings; therefore, this subset is used exclusively. This subset contains $409$ recordings from unique adult subjects, each annotated as Normal or Abnormal. Variable recording durations and inconsistent signal quality further increase the realism and challenge of the dataset.

\begin{table}[H]
\centering
\caption{Summary of the CinC dataset.}
\begin{tabular}{l l}
\hline
\textbf{Feature} & \textbf{Description} \\
\hline
Number of recordings & 409 (training-a subset) \\
Number of subjects & 409 \\
Classes & Normal / Abnormal \\
Age group & 18--85 years \\
Recording environment & Clinical / Non-clinical \\
Noise contamination & Moderate \\
\hline
\end{tabular}
\label{tab:cinC_summary}
\end{table}

\subsection{CirCor DigiScope Dataset 2022 (CirCor)}

The CirCor dataset, released for the George B. Moody PhysioNet Challenge 2022, represents one of the largest publicly available pediatric PCG collections. It includes $3,136$ recordings from $942$ subjects, gathered during two mass screening campaigns in northeastern Brazil, CC2014 and CC2015. Unlike CinC, which focuses on adults, CirCor is predominantly pediatric, with children constituting over $70\%$ of the cohort, complemented by infants, adolescents, and a small number of neonates.

Recordings were captured from up to five auscultation sites per subject, with the majority originating from the four canonical valve positions: aortic (AV), pulmonary (PV), tricuspid (TV), and mitral (MV). This multi-site structure enables spatial analysis of murmurs but introduces inter-recording variability for the same subject. The dataset provides two complementary annotation layers: ($1$) murmur status, categorized as Absent, Present, or Unknown, and ($2$) clinical outcome, labeled as Normal or Abnormal. For binary classification, this study uses only the Absent and Present murmur labels. CirCor presents several challenges inherent to real-world screening. Recordings contain pervasive environmental noise, including stethoscope handling artifacts, ambient speech, crying, and traffic sounds, some of which spectrally overlap with diagnostically relevant heart sounds. Additionally, the distribution of murmur labels is imbalanced, with significantly more Absent than Present cases, posing a challenge for supervised learning without careful handling of class representation.

\begin{table}[H]
\centering
\caption{Summary of the CirCor dataset.}
\begin{tabular}{l l}
\hline
\textbf{Feature} & \textbf{Description} \\
\hline
Number of subjects & 942 \\
Number of recordings & 3,136 \\
Classes (murmur) & Absent / Present (Unknown excluded) \\
Age group & Primarily children \\
Auscultation locations & AV, PV, TV, MV \\
Noise contamination & Moderate to high, real-world conditions \\
\hline
\end{tabular}
\label{tab:circor_summary}
\end{table}


While both datasets used in this study support binary heart sound classification, they differ fundamentally in scope and structure. The CinC dataset emphasizes adult cardiac assessment across heterogeneous acquisition conditions, with one recording per subject and a focus on global abnormality detection. In contrast, CirCor provides a pediatric-focused, multi-site recording structure with fine-grained murmur annotations and greater temporal and spatial variability.

\section{Methodology}

\subsection{Variational Convolution with Quantum-Inspired Sampling}

This research introduces QiVConv, a probabilistic convolutional layer that enhances uncertainty modeling through a novel sampling method. A demonstration implementation of the proposed methodology is publicly available on GitHub. The repository includes the QiVConv layer and the RFR block, \href{https://github.com/amingolnari/Demo-QiVC-Net}{https://github.com/amingolnari/Demo-QiVC-Net}.

Unlike standard variational layers that perturb kernel weights using unstructured Gaussian noise, QiVConv leverages structured stochasticity derived from principles of quantum state evolution. 
Specifically, a unitary-rotated variational sampling mechanism is introduced, in which structured uncertainty is injected via rotated variational sampling with subspace swapping, while preserving full differentiability and enabling more expressive posteriors.

The multichannel one-dimensional input signal is denoted by 
$\mathbf{X} \in \mathbb{R}^{T \times C_{\text{in}}}$, 
where $T$ represents the number of time steps and $C_{\text{in}}$ the number of input channels. 
The convolutional kernel 
$\mathbf{W} \in \mathbb{R}^{K \times C_{\text{in}} \times C_{\text{out}}}$, 
where $K$ is the kernel size and $C_{\text{out}}$ the number of filters, 
is modeled as a Gaussian-distributed parameter 
$\mathbf{W} \sim \mathcal{N}(\boldsymbol{\mu}, \boldsymbol{\sigma}^2)$, 
and samples during training are drawn using the re-parameterization trick proposed by Kingma \& Welling \cite{kingma2013auto}.

\begin{equation}
\mathbf{W} = \boldsymbol{\mu} + \boldsymbol{\sigma} \odot \boldsymbol{\epsilon}_{\text{init}}
\end{equation}

Here, $\boldsymbol{\mu}$ and $\boldsymbol{\sigma}$ represent the mean and standard deviation. The initial noise $\boldsymbol{\epsilon}_{\text{init}}$ is sampled from a Gaussian distribution. In the subsequent processing stage, this noise is transformed via QR decompositions, unitary rotations, and controlled depolarizing noise so that the resulting non-Gaussian perturbation $\boldsymbol{\epsilon}_{\text{QiRE}}$ replaces the conventional re-parameterization term $\boldsymbol{\sigma} \odot \boldsymbol{\epsilon}_{\text{init}}$.

\subsubsection{Quantum-Inspired Unitary-Rotated Variational Sampling}

Traditional variational inference typically injects randomness into the model weights using unstructured Gaussian noise. However, this approach can produce noise that is misaligned with the intrinsic geometry of the parameter space, potentially causing unstable training dynamics or poor calibration. To overcome this limitation, our method draws inspiration from the mathematical framework of quantum mechanics. Reliability-focused approaches have demonstrated that mathematically constrained, norm-preserving transformations within parameter spaces improve robustness to uncertainty \cite{lins2025quantum}. These insights align closely with the objectives of our QiVConv framework, which utilizes rotated variational sampling with subspace swap to inject structured stochasticity while maintaining geometric coherence.

\subsubsection{Inter-Channel Correlation and Kernel Vectorization}

A critical feature of the QiVConv mechanism is its holistic treatment of the kernel weights. The convolutional kernel $\mathbf{W}$ is first vectorized into a single vector $\mathbf{w} \in \mathbb{R}^{N}$, where $N = K \cdot C_{\text{in}} \cdot C_{\text{out}}$. By rotating the entire kernel vector $\mathbf{w}$ within its subspace, structured stochasticity is introduced across every component, including spatial and inter-channel correlations. This approach avoids the need for an explicit layer to model channel correlation, achieving an efficient coupled uncertainty injection over the entire parameter space.

In quantum theory, pure states are represented as unit vectors on a hypersphere within a complex Hilbert space, and their evolution is governed by unitary operations that rotate these vectors without altering their norm or internal structure \cite{sakurai2020modern}. This concept of geometry-preserving stochastic evolution offers a compelling analogy for constructing noise processes in neural networks. Accordingly, our approach treats the vectorized noise similarly to a quantum state: rather than injecting noise randomly, it undergoes a structured rotation within a low-dimensional subspace of the weight space, preserving both its norm and coherence.

The procedure to generate such noise is as follows:

\begin{enumerate}
\item \textbf{Flattening:} Reshape the kernel weight $\mathbf{W}$ (the noise $\boldsymbol{\epsilon}_{\text{init}}$) into a vector $\mathbf{w} \in \mathbb{R}^{N}$. 

\item \textbf{Random subspace generation:} Draw a random matrix $\mathbf{G} \in \mathbb{R}^{N \times k}$ with entries sampled from $\mathcal{N}(0, 1)$, and compute its QR decomposition to obtain an orthonormal basis:

\begin{equation}
\mathbf{G} = \mathbf{Q} \mathbf{R}, \quad \mathbf{Q} \in \mathbb{R}^{N \times k}.
\end{equation}

The columns of $\mathbf{Q}$ span a $k$-dimensional subspace of $\mathbb{R}^N$.

\item \textbf{Amplitude normalization:} Sample a base noise vector $\boldsymbol{\epsilon}_0 \sim \mathcal{N}(0, \mathbf{I}_N)$ and normalize it:

\begin{equation}
\boldsymbol{\epsilon} \leftarrow \frac{\boldsymbol{\epsilon}_0}{\|\boldsymbol{\epsilon}_0\|},
\end{equation}

projecting it onto the unit hypersphere, analogous to encoding a normalized quantum state.

\item \textbf{Unitary rotation:} Sample a rotation matrix $\mathbf{U} \sim \text{Haar}(\mathrm{SO}(k))$ and perform a structured rotation in the subspace defined by $\mathbf{Q}$:

\begin{equation}
\boldsymbol{\epsilon}_{\text{rot}} = \mathbf{Q}\mathbf{U}\mathbf{Q}^\top\boldsymbol{\epsilon}.
\end{equation}

This rotation whitens the noise and introduces structured stochasticity while preserving the norm and subspace structure.

\item \textbf{Variational update (subspace swap):} Construct the final noise vector by swapping the component of the initial noise that lies within the subspace $\mathbf{Q}$ with the newly rotated component $\boldsymbol{\epsilon}_{\text{rot}}$, while keeping the orthogonal component unchanged:

\begin{equation}
\boldsymbol{\epsilon}_{\text{final}} \leftarrow \boldsymbol{\epsilon} - \mathbf{Q}\mathbf{Q}^\top\boldsymbol{\epsilon} + \boldsymbol{\epsilon}_{\text{rot}}.
\end{equation}

This operation ensures that the noise remains aligned with the rotational subspace while preserving the overall variance of the original noise.

\item \textbf{Optional decoherence:} Optionally apply depolarizing noise to simulate decoherence effects:

\begin{equation}
\boldsymbol{\epsilon}_{\text{final}} \leftarrow \mathbf{m} \odot \boldsymbol{\epsilon}_{\text{final}} + (1 - \mathbf{m}) \cdot \frac{\mathbf{1}}{\sqrt{N}}, \quad \mathbf{m} \sim \text{Bernoulli}(1 - p),
\end{equation}

\noindent where $p$ controls the noise level and $\mathbf{1}$ is the unit vector.
\end{enumerate}

The entire convolutional kernel $\mathbf{W}$ is flattened into a single global vector $\mathbf{w}$, treating the full layer as a unified geometric space. This whole-layer vectorization couples all filter and channel parameters together, allowing the Haar rotation matrix to induce complex-valued, structured cross-correlations across channel and filter boundaries during the unitary transformation phase. To isolate orientation modifications from scale alterations, the raw noise vector is amplitude-normalized to unit length. Crucially, any potential parameter-dilution effects caused by high values of $N$ are adaptively countered downstream by the element-wise learnable standard deviation parameter, which scales the structured perturbation locally for each weight element. 

The resulting noise vector $\boldsymbol{\epsilon}_{\text{final}}$ is reshaped to match the dimensions of $\mathbf{W}$ and applied in the variational convolution. The final weights then compute as $\mathbf{W} = \boldsymbol{\mu} + \boldsymbol{\sigma} \odot \boldsymbol{\epsilon}_{\text{final}}$. It was experimentally observed that setting the subspace dimensionality $k$ between $2$ and $9$ provides computational efficiency while maintaining performance. In the experiments conducted, the value of $k$ is set to $5$. This unitary-rotated variational sampling introduces coherent geometric structures into the variational space, leading to smoother uncertainty propagation, enhanced regularization, and more calibrated predictive distributions. The overall workflow is outlined in Algorithm \ref{alg:QiVC}. To clarify for the readers, we define $\epsilon_0$ as the initial Gaussian noise, $\epsilon_{\text{rot}}$ as the rotated component, and $\epsilon_{\text{final}}$ as the final structured noise, while explicitly annotating all matrix dimensions and operations for transparency. 

To satisfy unbiased optimization, the random subspace basis matrix $Q$ and the rotation matrix $R$ are resampled at every single forward pass during training. Conversely, during validation and testing, the sampling routine is eliminated, ensuring a fully deterministic inference pass based on the trained posterior mean $\mu$.

\begin{algorithm}[h!]
\caption{QiVC layer: quantum-inspired variational convolution}
\label{alg:QiVC}
\begin{algorithmic}[1]
\Require Kernel $\mathbf{W}$ (random), subspace dim. $k = 5$, decoherence $p = 0.01$, prior var. $\sigma_{\text{prior}}^2 = 0.01$, KL scale $\lambda = 1\times10^{-5}$, noise $\boldsymbol{\epsilon}_0$
\Ensure Noisy kernel $\mathbf{W}_{\text{noisy}}$, KL loss

\Function{OrthonormalBasis}{$N,k$}
 \State $\mathbf{G}\!\sim\!\mathcal{N}(0,1)^{N\times k}$ \Comment{Generate random Gaussian matrix of size $N \times k$}
 \State $[\mathbf{Q},\mathbf{R}]\!\leftarrow\!\mathrm{QR}(\mathbf{G})$ \Comment{Compute QR decomposition for orthonormal basis}
 \State \Return $\mathbf{Q}$ \Comment{Return orthonormal basis of the subspace}
\EndFunction

\Function{HaarSO}{$k$}
 \State $\mathbf{A}\!\sim\!\mathcal{N}(0,1)^{k\times k}$ \Comment{Random Gaussian matrix for generating Haar unitary}
 \State $[\mathbf{Q},\mathbf{R}]\!\leftarrow\!\mathrm{QR}(\mathbf{A})$ \Comment{Compute QR decomposition to orthogonalize}
 \State $\mathbf{D}\!\leftarrow\!\operatorname{diag}(\operatorname{sign}(\operatorname{diag}(\mathbf{R})))$ \Comment{Correct signs for a proper orthogonal matrix}
 \State $\mathbf{U}\!\leftarrow\!\mathbf{Q}\mathbf{D}$ \Comment{Unitary matrix sampled from Haar measure}
 
 \If{$\det(\mathbf{U})<0$} 
  \State $\mathbf{U}_{:,1}\!\leftarrow\!-\mathbf{U}_{:,1}$ \Comment{Ensure special orthogonal matrix (det=+1)}
 \EndIf
 \State \Return $\mathbf{U}$ \Comment{Return rotation matrix}
\EndFunction

\Function{QiVCSample}{$\mathbf{W},k,p$}
 \State $\mathbf{w}\!\leftarrow\!\operatorname{vec}(\mathbf{W})$ \Comment{Flatten kernel to vector}
 \State $N\!\leftarrow\!|\mathbf{w}|$ \Comment{Compute Number of elements in kernel}
 \State $\boldsymbol{\epsilon}\!\sim\!\mathcal{N}(0,\mathbf{I}_N)$ \Comment{Sample isotropic Gaussian noise}
 \State $\boldsymbol{\epsilon}\!\leftarrow\!\boldsymbol{\epsilon}_0/\|\boldsymbol{\epsilon}_0\|_2$ \Comment{Normalize noise vector (amplitude normalization)}
 \State $\mathbf{Q}\!\leftarrow$\Call{OrthonormalBasis}{$N,k$} \Comment{Sample random subspace of dimension $k$}
 \State $\mathbf{U}\!\leftarrow$\Call{HaarSO}{$k$} \Comment{Sample random unitary rotation in subspace}
 \State $\boldsymbol{\epsilon}_{\text{rot}}\!\leftarrow\!\mathbf{Q}\mathbf{U}\mathbf{Q}^\top\boldsymbol{\epsilon}$ \Comment{Rotate noise in low-dimensional subspace}
 \State $\boldsymbol{\epsilon}_{\text{final}}\!\leftarrow\!\boldsymbol{\epsilon}-\mathbf{Q}\mathbf{Q}^\top\boldsymbol{\epsilon}+\boldsymbol{\epsilon}_{\text{rot}}$ \Comment{Combine rotated and orthogonal noise components}
 \If{$p>0$} 
  \State $\mathbf{m}\!\sim\!\mathrm{Bernoulli}(1-p)^N$ \Comment{Sample decoherence mask (drop probability $p$)}
  \State $\boldsymbol{\epsilon}_{\text{final}}\!\leftarrow\!\mathbf{m}\odot\boldsymbol{\epsilon}_{\text{final}}+(1-\mathbf{m})/\sqrt{N}$ \Comment{Apply decoherence to noise}
 \EndIf
 \State \Return $\operatorname{reshape}(\boldsymbol{\epsilon}_{\text{final}},\operatorname{shape}(\mathbf{W}))$ \Comment{Reshape noise to original kernel shape}
\EndFunction

\Procedure{QiVCLayer}{$\mathbf{X},\boldsymbol{\mu},\boldsymbol{\rho},\mathbf{b},k,p,\lambda,\sigma_{\text{prior}}$}
 \State $\boldsymbol{\sigma}\!\leftarrow\!\operatorname{softplus}(\boldsymbol{\rho})$ \Comment{Compute standard deviation via softplus to ensure positivity}
 \State $\mathbf{W}_s\!\leftarrow\!\boldsymbol{\mu}+\boldsymbol{\sigma}\odot$\Call{QiVCSample}{$\boldsymbol{\mu},k,p$} \Comment{Sample noisy kernel using reparameterization trick}
 \State $\mathbf{Z}\!\leftarrow\!\phi(\mathcal{C}(\mathbf{X};\mathbf{W}_s)+\mathbf{b})$ \Comment{Forward pass: convolution + bias + activation}
 \State $\mathcal{L}\!\leftarrow\!\text{TaskLoss}(\mathbf{Z},\mathbf{Y})+\lambda\!\sum\!\left[\tfrac{\boldsymbol{\sigma}^2+\boldsymbol{\mu}^2}{2\boldsymbol{\sigma}_{\text{prior}}^2}-\log\boldsymbol{\sigma}+\log\boldsymbol{\sigma}_{\text{prior}}-\tfrac{1}{2}\right]$ \Comment{Total loss = task loss + KL divergence}
 \State \Return $\mathcal{L}$, $\mathbf{Z}$ \Comment{Return loss and layer output}
\EndProcedure
\end{algorithmic}
\end{algorithm}

\subsubsection{Variational Inference and Regularization}

The total objective of the QiVConv layer is defined by minimizing the evidence lower bound (ELBO), as the result of combining the task-specific loss with a KL divergence term that regularizes the weight distributions. The variational posterior is parameterized by a mean $\boldsymbol{\mu}$ and a standard deviation $\boldsymbol{\sigma}$. The standard deviation is derived from a learnable parameter $\boldsymbol{\rho}$ via the softplus function to enforce non-negativity:
$$
\boldsymbol{\sigma} = \text{softplus}(\boldsymbol{\rho}) = \log(1 + e^{\boldsymbol{\rho}}).
$$

During inference (test time), the deterministic output $\hat{\mathbf{Z}}$ is computed by using the posterior mean of the variational kernel, which is defined as:

\begin{equation}
\hat{\mathbf{Z}} = \phi\left(
\mathcal{C}(\mathbf{X}; \boldsymbol{\mu}) + \mathbf{b} \right),
\end{equation}

\noindent where $\mathcal{C}(\cdot)$ denotes the convolution operator, $\mathbf{b}$ is the bias term, and $\phi$ is a non-linear activation function. To regularize the variational distribution, the Kullback–Leibler (KL) divergence between the learned posterior distribution $q(\mathbf{W}|\boldsymbol{\mu}, \boldsymbol{\sigma})$ and the prior distribution $p(\mathbf{W}|\sigma_{\text{prior}})$ is minimized:

\begin{equation}
\mathrm{KL}(q \|\, p) = \sum \left[
\frac{\boldsymbol{\sigma}^2 + \boldsymbol{\mu}^2}{2\boldsymbol{\sigma}_\text{prior}^2}
- \log(\boldsymbol{\sigma} + \varepsilon)
+ \log \boldsymbol{\sigma}_\text{prior}
- \frac{1}{2}
\right],
\end{equation}

The sum is computed over all weights and biases in the convolutional layer, and the small constant $\varepsilon$ ensures numerical stability. The total objective function minimized during training is the following:

\begin{equation}
\mathcal{L}_{\text{Total}} = \mathcal{L}_{\text{Task}}(\mathbf{Z}, \mathbf{Y}) + \lambda \cdot \mathrm{KL}(q \|\, p),
\end{equation}

\noindent where $\mathcal{L}_{\text{Task}}$ is the data likelihood loss and $\lambda$ is a weight factor controlling the strength of the $KL$ regularization. Since the perturbation path incorporates structural constraints (unit-norm bounds and low-dimensional projections), the true density of the transformed noise vector slightly deviates from an unconstrained isotropic Gaussian. Consequently, the $KL(q \|\, p)$ should be interpreted as a well-behaved structured variational regularizer within an implicit variational framework, rather than a strictly analytical closed-form $KL$ penalty for non-Gaussian fields. This term effectively guides the high-dimensional parameter configurations toward a stable, physically meaningful Gaussian prior.

\subsubsection{Connection to Structured Variational Inference}

Unlike standard mean-field variational inference, which assumes independent Gaussian distributions over model parameters, the proposed QiRE mechanism introduces structured perturbations that induce correlations among weights. This is achieved by projecting noise into a low-dimensional subspace and applying orthogonal transformations, resulting in a geometry-preserving perturbation process. From a probabilistic perspective, this can be interpreted as an implicit approximation to a correlated posterior distribution, where dependencies are induced through structured sampling rather than explicitly parameterized covariance matrices. This places the proposed method within the broader class of structured and subspace-based variational inference approaches.

Beside this, QiVC-Net primarily models epistemic uncertainty (parameter uncertainty) through variational weight distributions. This is distinct from aleatoric uncertainty (data uncertainty), for which the uncertainty can be hardly reduced. For PCG classification, epistemic uncertainty is particularly relevant: it captures model confidence regarding inter-subject variability and recording conditions that may differ from the training distribution.

\subsection{Pre-processing}

The preprocessing pipeline was designed to enhance the quality and consistency of PCG recordings while preserving diagnostically relevant acoustic features. All recordings were band-pass filtered between $\SI{25}{Hz}$ and $\SI{400}{Hz}$ using a fourth-order zero-phase Butterworth filter, consistent with established PCG signal processing practices reported in previous studies \cite{kalatehjari2025advanced}. This frequency range captures the main components of S$_1$ and S$_2$ heart sounds and most pathological murmurs, while attenuating respiratory noise and high-frequency artifacts.

Each recording was then segmented into consecutive, non-overlapping windows of fixed $\SI{4}{s}$ duration. This approach ensures complete temporal coverage and uniform segment length suitable for batch processing. Each segment contains multiple cardiac cycles across typical heart rates ($60$ to $150$ bpm), allowing the model to capture continuous acoustic dynamics, including murmurs, pauses, and inter-cycle variability, without assuming regular cardiac periodicity.

To guarantee rigorous validation integrity and prevent data leakage on the CirCor dataset, the necessary patient partitioning is executed strictly at the subject level prior to any audio segmentation. In our design we select a single representative recording per patient before window slicing, ensuring subject-level independence. Specifically, we leverage the \quotes{most audible location} annotation provided in the dataset metadata to identify the recording with the highest diagnostic yield for each subject. This strategy is grounded in the findings of Elola et al. \cite{elola2023beyond}, who demonstrated that murmur intensity and detectability vary significantly across auscultation sites, and that focusing on the most informative location is critical for robust automated grading and classification. By isolating this single, clinically significant recording per patient, we ensure that all subsequent segments originate from a unique source.

\begin{figure}[H]
\begin{center}
\includegraphics[width = 14 cm, clip = true, trim = 0 0 0 0]{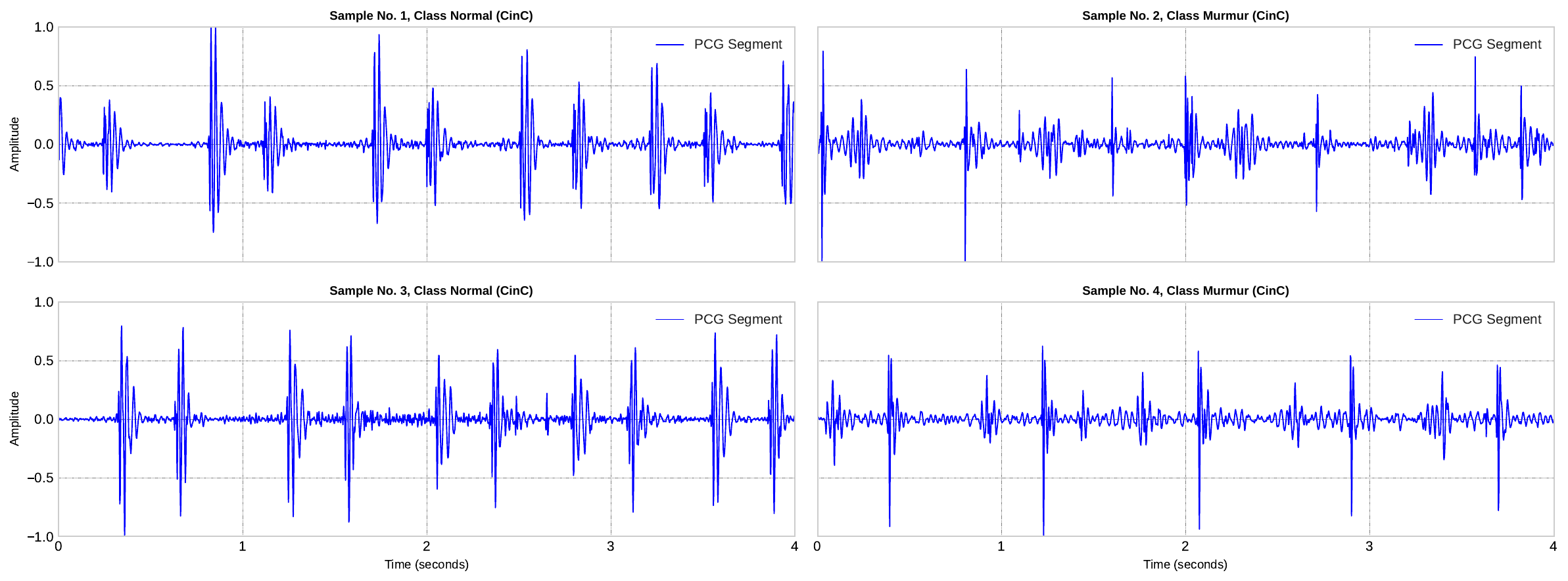}
\caption{Representative PCG signal segments from the CinC dataset, depicting normal and with murmur heart sounds across different recording conditions.}
\label{fig:cinc_samples}
\end{center}
\end{figure}

\begin{figure}[H]
\begin{center}
\includegraphics[width = 14 cm, clip = true, trim = 0 0 0 0]{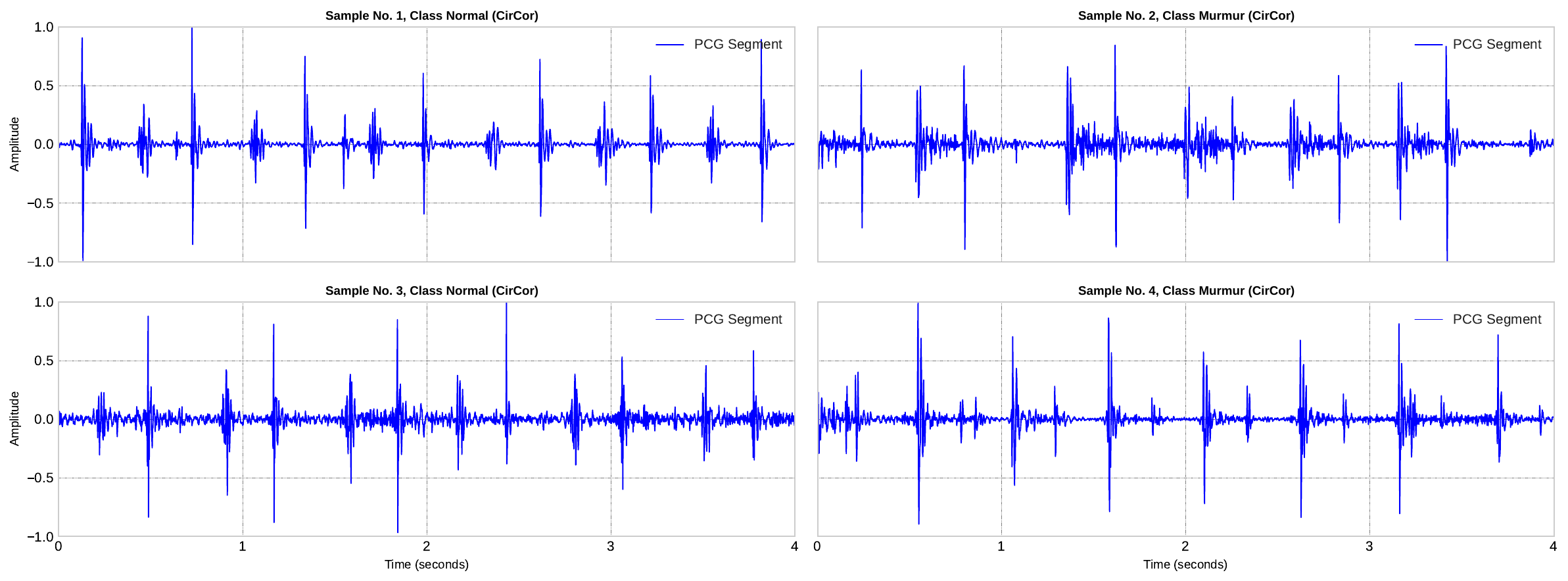}
\caption{Representative PCG signal segments from the CirCor dataset, depicting normal and with murmur heart sounds across different recording conditions.}
\label{fig:circor_samples}
\end{center}
\end{figure}

Finally, each segment was resampled to $2000$ samples (effective $\SI{500}{Hz}$), mean-centered to remove baseline drift, and amplitude-normalized to the range $\pm1.0$. Segments containing non-finite or zero-valued samples were discarded to maintain numerical stability during training. Figures \ref{fig:cinc_samples} and \ref{fig:circor_samples} present representative PCG signal samples from the CinC and CirCor datasets, respectively, illustrating the distinct characteristics and variability of heart sounds across both sources.

\subsection{Evaluation Metrics}

To evaluate classification models, several key metrics are used to assess accuracy, robustness, and clinical reliability. These are derived from the confusion matrix, which defines true positives (TP), false positives (FP), true negatives (TN), and false negatives (FN), with abnormal heart sounds treated as positive and normal sounds as negative. Accuracy measures the overall proportion of correct predictions. Sensitivity (recall) evaluates the model's ability to detect abnormal cases, which is critical in medical diagnosis. Specificity measures correct identification of normal cases, reducing false alarms. The F1 score provides a unified measure of model performance by harmonizing true positive rate and positive predictive value, making it particularly valuable in scenarios where class distributions are skewed. Together, these metrics provide a comprehensive evaluation of a model’s performance in clinical and real-world settings.

The mathematical formulations of these evaluation metrics are as follows. Accuracy is defined as the ratio of correctly predicted instances to the total number of instances:

\begin{equation}
\text{Accuracy} = \frac{TP + TN}{TP + TN + FP + FN}
\end{equation}

\begin{equation}
\text{Sensitivity} = \frac{TP}{TP + FN}
\end{equation}

\begin{equation}
\text{Specificity} = \frac{TN}{TN + FP}
\end{equation}

\begin{equation}
\text{F1-score} = \frac{2 \cdot TP}{2 \cdot TP + FP + FN}
\end{equation}

These formulas collectively support a rigorous and clinically meaningful evaluation of PCG classification approaches, enabling reliable comparison across different models and datasets.

\subsection{Proposed Architecture}

In this research, a novel deep learning architecture for PCG classification is proposed, integrating temporal symmetry modeling, uncertainty-aware representation learning, and hierarchical feature extraction. The model is built upon the RFR block, a custom residual unit designed to capture bidirectional physiological dynamics in heart sound sequences. Each RFR block processes the input through a dual-path scheme consisting of a forward path and a time-reversed backward path, both operating on the same segment. The outputs of these paths are then fused using long short-term memory (LSTM) layers and residual connections, enabling the model to capture both local and long-range temporal dependencies. Instead of employing a simple addition for the residual connection, the shortcut is concatenated with the fused features and the result is refined using an LSTM layer. The RFR block consists of three main components:

\begin{enumerate}
    \item \textbf{Shortcut path:} A 1$\times$1 convolution followed by batch normalization and a non-linear activation provides a residual connection, refine channel dimension without learning entirely new representations.
    
    \item \textbf{Forward and backward paths:} The forward path applies a QiVConv layer followed by batch normalization and activation. The backward path processes a time-reversed version of the input through an identical QiVConv layer and then reverses the output back to the original temporal order. This design captures complementary forward and reverse temporal features.
    
    \item \textbf{Fusion and refinement:} The outputs of the forward and backward paths are concatenated and fused using a LSTM layer. Batch normalization and activation follow to produce a refined representation. Finally, the fused features are concatenated with the shortcut path and further processed by another LSTM, batch normalization, and activation to produce the final block output.
\end{enumerate}

\begin{figure}[H]
\begin{center}
\includegraphics[width = 12 cm, clip = true, trim = 100 100 80 110]{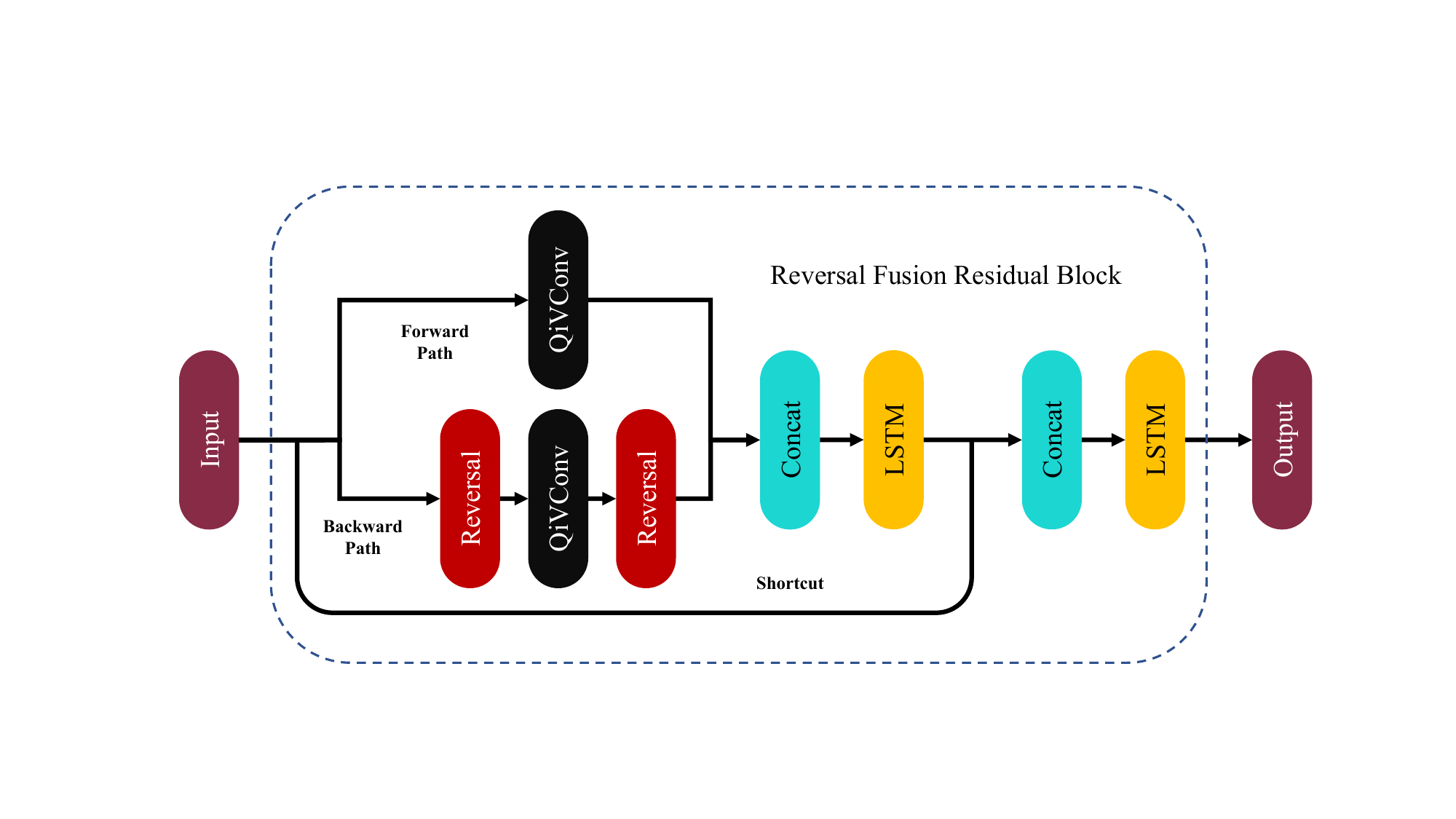}
\caption{Reversal fusion residual block architecture.}
\label{fig:rfr_block}
\end{center}
\end{figure}

Multiple RFR blocks are stacked with progressively increasing filter counts and interspersed downsampling, enabling hierarchical abstraction of PCG features. Global max pooling is applied before the output layer to ensure invariance to sequence shifts. Finally, a fully connected layer followed by a softmax classifier produces a binary decision between normal and abnormal classes. The RFR block architecture is illustrated in Figure \ref{fig:rfr_block}.

\subsection{Training Strategy}

\begin{figure}[H]
\centering
\begin{subfigure}{0.49\textwidth}
    \centering
    \includegraphics[width=0.8\linewidth, clip=true, trim=0 0 0 0]{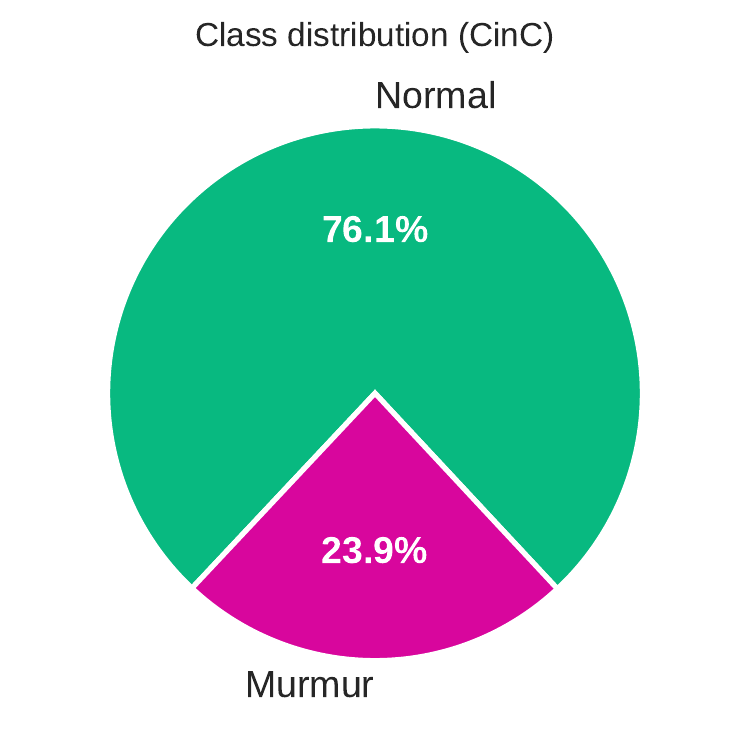}
    \caption{}
\end{subfigure}
\hfill
\begin{subfigure}{0.49\textwidth}
    \centering
    \includegraphics[width=0.8\linewidth, clip=true, trim=0 0 0 0]{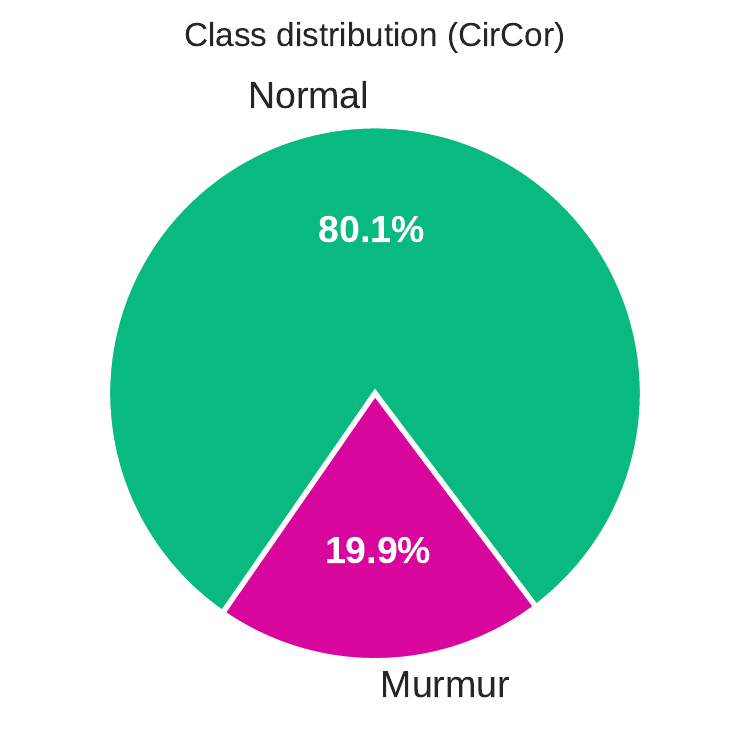}
    \caption{}
\end{subfigure}
\caption{Class distribution in the PCG datasets: (a) CinC and (b) CirCor.}
\label{fig:cinc_circor_pie}
\end{figure}

As shown in Figure \ref{fig:cinc_circor_pie}, there is a significant class imbalance in both the CinC and CirCor datasets, with a larger proportion of normal heart sound recordings compared to abnormal ones. To effectively address this issue, the proposed QiVC-Net is trained using a composite loss function that combines categorical cross-entropy (CCE) denoted as $\mathcal{L}_{\text{CCE}}$ and Dice loss denoted as $\mathcal{L}_{\text{Dice}}$, balancing sample-wise classification accuracy with region-level overlap. A dynamic loss weighting mechanism, inspired by our previous work \cite{golnari2026dmf}, adaptively penalizes and reweights each loss component in real time during training, allowing the model to emphasize the most informative objective at different stages. This adaptive strategy improves convergence and generalization under noisy and data-imbalanced conditions. 

Within the QiVConv layers, the structured noise generation follows the default configuration outlined in Algorithm \ref{alg:QiVC}: a subspace dimensionality of $k=5$ was selected to balance computational efficiency with expressive capacity, and a depolarization rate of $p=0.01$ was applied to simulate mild decoherence effects. The variational posterior was regularized against a unit Gaussian prior ($\sigma_{\text{prior}}=1.0$) with a KL-divergence scaling factor of $\lambda=1 \times 10^{-5}$. Additionally, to prevent overfitting, early stopping was applied alongside model checkpointing to retain the best-performing weights based on validation performance throughout training.

\begin{equation}
\mathcal{L}_{\text{CCE}} = -\frac{1}{N} \sum_{i=1}^{N} \sum_{c=1}^{C} y_{i,c} \log(\hat{y}_{i,c})
\label{eq:cce}
\end{equation}

\begin{equation}
\mathcal{L}_{\text{Dice}} = 1 - \frac{2 \sum_{i=1}^{N} y_i \hat{y}_i}{\sum_{i=1}^{N} y_i + \sum_{i=1}^{N} \hat{y}_i}
\label{eq:dice}
\end{equation}

The training process was conducted for over $500$ epochs per experiment, with a batch size of $256$ and a learning rate of $1\times10^{-3}$, using stratified $5$-fold cross-validation to ensure balanced representation of both classes in each fold. Additionally, to prevent overfitting, early stopping was employed alongside model checkpointing, ensuring that the training process is terminated once the performance on the validation set ceased to improve. In this strategy, the model continuously monitors the performance at each epoch and stores the weights corresponding to the best observed validation metric. This approach ensures that, even if overfitting occurs in later epochs, the model retains the set of weights that achieved the best generalization performance on the validation set for final evaluation and inference \cite{golnari2024probabilistic}. Experimental observations revealed that the optimal validation performance does not always coincide with the minimum training loss, as excessive loss minimization may lead to overfitting or reduced generalization capability.

\section{Results}

This section presents the experimental evaluation of the proposed QiVC-Net on two benchmark datasets for PCG classification. To ensure robustness and statistical reliability, we performed $5$-fold stratified cross-validation on both datasets and report key performance metrics including accuracy, sensitivity, specificity, and F1-score. The following results highlight the model’s ability to distinguish between normal and abnormal cardiac auscultations under real-world conditions characterized by noise, signal variability, and significant class imbalance.

Table \ref{tab:kfold_results_combined} presents the $5$-fold stratified cross-validation results of the proposed QiVC-Net on the CinC and CirCor datasets. As shown in the table, the model achieves strong and consistent performance across both datasets. On the CinC dataset, it records an average accuracy of $97.84\%$, with specificity, sensitivity, and F1-score of $98.14\%$, $96.89\%$, and $97.41\%$, respectively, demonstrating balanced classification capability. Similarly, on the CirCor dataset, the model attains an average accuracy of $97.89\%$, a high specificity of $99.27\%$, a sensitivity of $92.34\%$, and an F1-score of $95.71\%$. These results highlight the model’s strong generalization across folds, reliable detection of both normal and abnormal heart sounds, and robustness to class imbalance.

\begin{table}[H]
\centering
\small
\caption{5-Fold stratified cross-validation results on CinC and CirCor datasets.}
\begin{tabular}{l l c c c c}
\hline
\textbf{Dataset} & \textbf{Fold} & \textbf{Accuracy (\%)} & \textbf{Specificity (\%)} & \textbf{Sensitivity (\%)} & \textbf{F1-score (\%)} \\
\hline
\multirow{6}{*}{\textbf{CinC}} 
 & Fold 1 & 98.44 & 98.59 & 97.98 & 96.73 \\
 & Fold 2 & 96.57 & 96.13 & 97.89 & 93.45 \\
 & Fold 3 & 98.87 & 99.04 & 98.28 & 97.56 \\
 & Fold 4 & 97.99 & 99.41 & 93.66 & 95.82 \\
 & Fold 5 & 97.31 & 97.51 & 96.65 & 94.43 \\
 & Average & 97.84 & 98.14 & 96.89 & 95.60 \\
\hline
\multirow{6}{*}{\textbf{CirCor}} 
 & Fold 1 & 98.64 & 99.54 & 94.98 & 96.50 \\
 & Fold 2 & 98.28 & 99.61 & 93.28 & 95.79 \\
 & Fold 3 & 98.14 & 99.48 & 92.52 & 95.03 \\
 & Fold 4 & 95.96 & 98.14 & 87.11 & 89.48 \\
 & Fold 5 & 98.42 & 99.58 & 93.78 & 95.95 \\
 & Average & 97.89 & 99.27 & 92.34 & 95.55 \\
\hline
\end{tabular}
\label{tab:kfold_results_combined}
\end{table}

\begin{table}[H]
\centering
\caption{Benchmarking of various deep learning-based PCG classification methods on the CirC and CirCor datasets. “NA” indicates that the corresponding metric was not reported in the original study.}
\label{tab:pcg_comparison}
\begin{adjustbox}{max width=\textwidth}
\begin{tabular}{l M{0.65cm} M{7cm} M{1.65cm} M{1.65cm} M{1.65cm} M{1.65cm}}
\hline
\textbf{Dataset} & \textbf{Paper} & \textbf{Methodology} & \textbf{Accuracy (\%)} & \textbf{Sensitivity (\%)} & \textbf{Specificity (\%)} & \textbf{F1-score (\%)} \\
\hline
\multirow{11}{*}{CirC} 
 & \cite{hsieh2025development} & Recognizing recurrent neural network (RRNN) & 95.20 & 91.60 & 99.10 & NA \\
 & \cite{patwa2025heart} & 1D-CNN, wavelet stockwell transform (WST) & 96.51 & 96.60 & 96.60 & 96.42 \\
 & \cite{li2020classification} & Hidden semi-Markov model (HSMM), CNN & 86.80 & 87.00 & 86.60 & NA \\
 & \cite{dastagir2021computer} & Support vector machine (SVM) & 91.36 & 80.28 & 94.47 & NA \\
 & \cite{chen2022automatic} & CNN-LSTM & 86.00 & 87.00 & 89.00 & 91.00 \\
 & \cite{khan2020automatic} & Ensemble classifier (SVM, KNN, DT, ANN, LSTM) & 91.23 & 78.81 & 97.04 & NA \\
 & \cite{krishnan2020automated} & Feed-forward neural network (FNN) & 85.65 & 86.73 & 84.75 & 84.58 \\
 & \cite{ghosh2022automated} & Stacked autoencoder deep neural network (SAE-DNN) & 95.43 & 97.92 & 98.32 & 95.75 \\
 & \cite{eneriz2024low} & 1D U-Net-based & 92.80 & 94.50 & NA & NA \\
 & \cite{soares2020autonomous} & Zero-order autonomous learning multiple-model (ALMMo-0) & 93.04 & 90.82 & 95.26 & NA \\
 & \textbf{Ours} & \textbf{QiVC-Net} & \textbf{97.84} & \textbf{96.89} & \textbf{98.14} & \textbf{95.60} \\

\hline
\multirow{7}{*}{CirCor} 
 & \cite{yen2025development} & SVM & 82.00 & NA & NA & NA \\
 & \cite{manshadi2024murmur} & Stockwell transform, AlexNet deep features, random forest (RF) & 92.10 & 91.00 & 91.00 & NA \\
 & \cite{vimalajeewa2025multiscale} & SVM & 76.61 & 82.12 & 54.03 & NA \\
 & \cite{zhang2024intelligent} & Parallel-attentive transformer & 79.80 & NA & NA & 65.10 \\
 & \cite{nguyen2024heart} & ResNet-101 & 86.00 & 86.00 & NA & 86.00 \\
 & \cite{fuadah2022optimal} & MFCC, KNN & 76.31 & NA & NA & NA \\
 & \cite{eneriz2024low} & 1D U-Net-based & 90.30 & 96.20 & NA & NA \\
 & \cite{patwa2025heart} & 1D-CNN, Wavelet stockwell transform (WST) & 90.09 & 90.14 & 90.14 & 90.09 \\
 & \textbf{Ours} & \textbf{QiVC-Net} & \textbf{97.89} & \textbf{92.34} & \textbf{99.27} & \textbf{95.55} \\

\hline
\end{tabular}
\end{adjustbox}
\end{table}

Table \ref{tab:pcg_comparison} summarizes the results of key PCG classification methods, including CNN-, RNN-, Transformer-based, and hybrid models. As shown in Table \ref{tab:pcg_comparison}, while many previous methods achieve high specificity, sensitivity is often lower due to class imbalance and noisy signals. In contrast, the proposed QiVC-Net demonstrates consistently strong and balanced performance across both datasets. On the CinC dataset, QiVC-Net achieves an average accuracy of $97.84\%$, specificity of $98.14\%$, sensitivity of $96.89\%$, and F1-score of $95.60\%$, whereas on the CirCor dataset it reaches an average accuracy of $97.89\%$, specificity of $99.27\%$, sensitivity of $92.34\%$, and F1-score of $95.55\%$. These results highlight that QiVC-Net not only maintains high overall classification performance but also effectively handles both normal and abnormal heart sounds, demonstrating robustness and stability across folds and datasets in the presence of noise and class imbalance.

\begin{figure}[H]
\centering
\begin{subfigure}{0.49\textwidth}
    \centering
    \includegraphics[width=\linewidth, clip=true, trim=0 0 0 0]{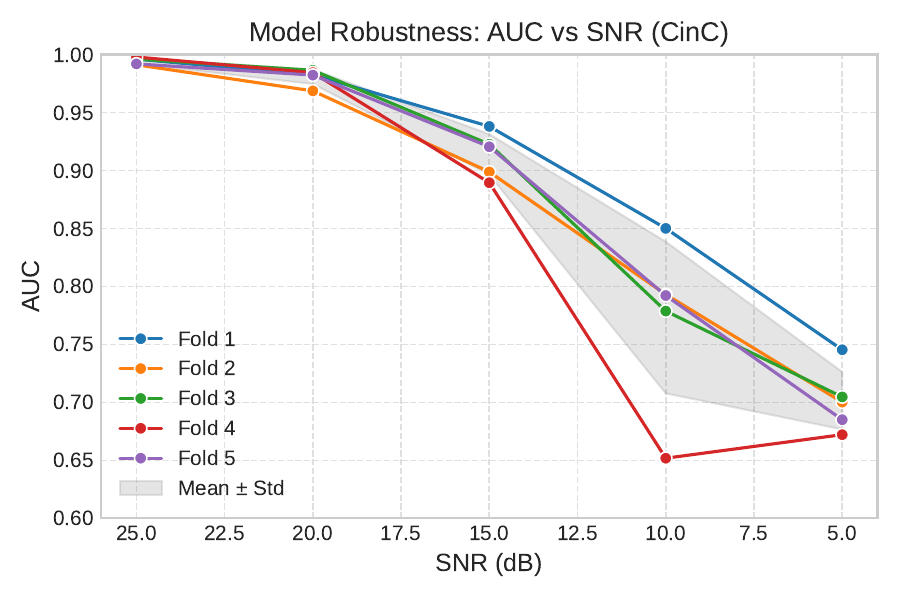}
    \caption{}
\end{subfigure}
\hfill
\begin{subfigure}{0.49\textwidth}
    \centering
    \includegraphics[width=\linewidth, clip=true, trim=0 0 0 0]{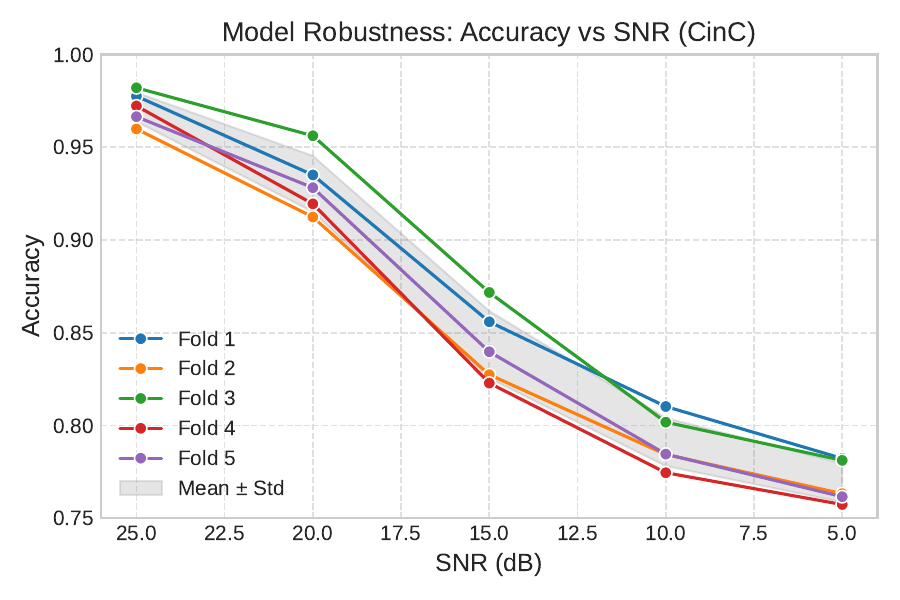}
    \caption{}
\end{subfigure}
\caption{Model robustness on the CinC dataset under varying SNR ratios; (a) AUC vs. SNR, and (b) Accuracy vs. SNR.}
\label{fig:cinc_snr}
\end{figure}

\begin{figure}[H]
\centering
\begin{subfigure}{0.49\textwidth}
    \centering
    \includegraphics[width=\linewidth, clip=true, trim=0 0 0 0]{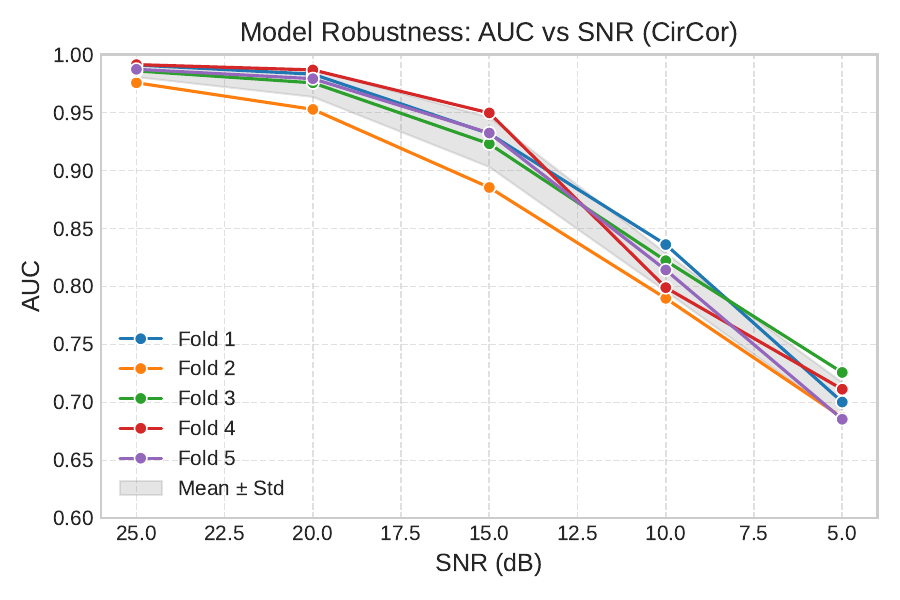}
    \caption{}
\end{subfigure}
\hfill
\begin{subfigure}{0.49\textwidth}
    \centering
    \includegraphics[width=\linewidth, clip=true, trim=0 0 0 0]{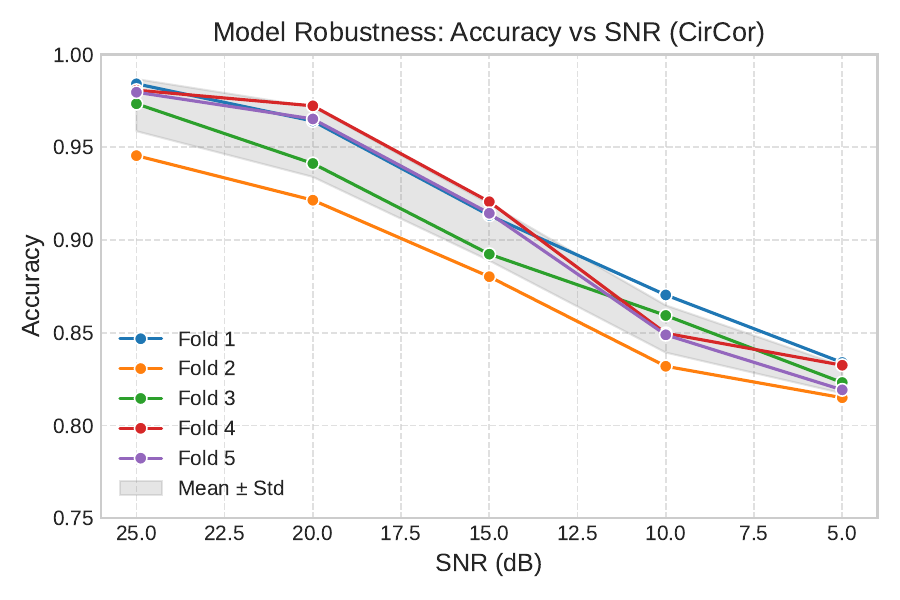}
    \caption{}
\end{subfigure}
\caption{Model robustness on the CirCor dataset under varying SNR ratios; (a) AUC vs. SNR, and (b) Accuracy vs. SNR.}
\label{fig:circor_snr}
\end{figure}

Figures \ref{fig:cinc_snr} and \ref{fig:circor_snr} illustrate the robustness of the proposed QiVC-Net against additive noise on the CinC and CirCor datasets, respectively. In these figures, the x-axis represents the signal-to-noise ratio (SNR) in dB displayed in descending order ($25 \rightarrow 5$ dB) to visually emphasize performance degradation as noise intensity increases from left to right, while the y-axis reports the area under the curve (AUC) metric quantifying classification discriminability. As shown in the figures across both datasets, the model maintains high classification performance at elevated SNR levels, achieving near-optimal AUC and accuracy, and exhibits a smooth, gradual degradation as noise intensity increases. On the CinC dataset, for example, the model achieves an average AUC around $0.99$ and accuracy above $0.97$ at $25$ dB SNR, while at $10$ dB and $5$ dB SNRs, accuracy gracefully declines to approximately $0.80$ and $0.76$. Similarly, on the CirCor dataset, the model retains accuracies above $0.96$ at high SNRs and maintains competitive performance around $0.85$ and $0.82$ at lower SNRs, with AUC showing a controlled decrease under severe noise. This consistent behavior across datasets indicates that QiVC-Net preserves discriminative capability and confidence calibration even in noisy and challenging recording conditions. The results highlight the robustness and generalization of the variational, uncertainty-aware architecture, demonstrating its suitability for real-world PCG analysis where signal noise is inevitable.

\begin{figure}[H]
\centering
\begin{subfigure}{0.49\textwidth}
    \centering
    \includegraphics[width=\linewidth, clip=true, trim=0 0 0 0]{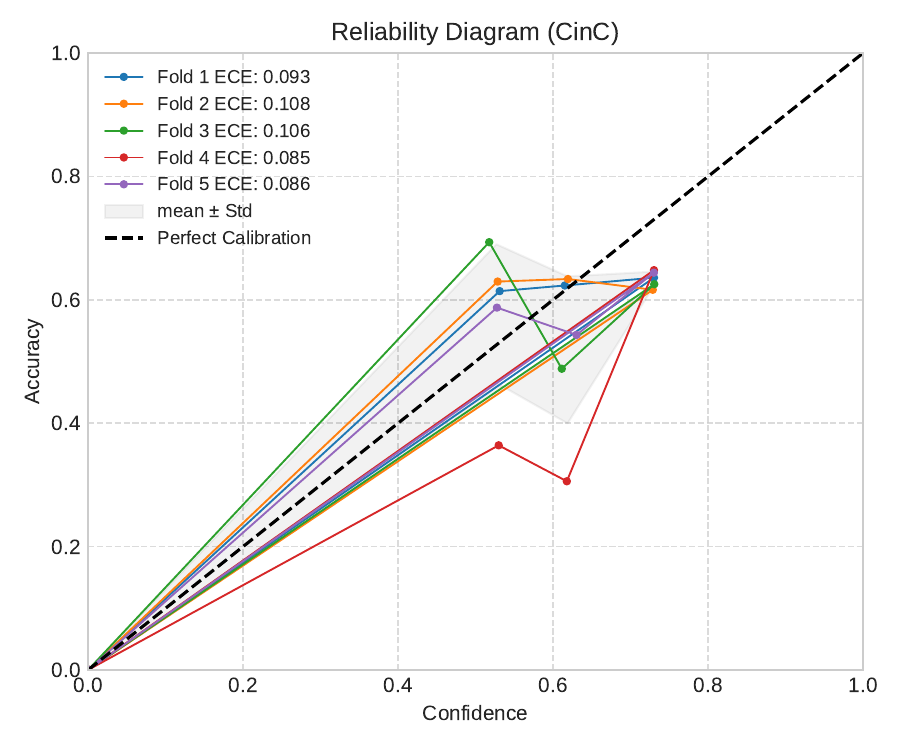}
    \caption{}
\end{subfigure}
\hfill
\begin{subfigure}{0.49\textwidth}
    \centering
    \includegraphics[width=\linewidth, clip=true, trim=0 0 0 0]{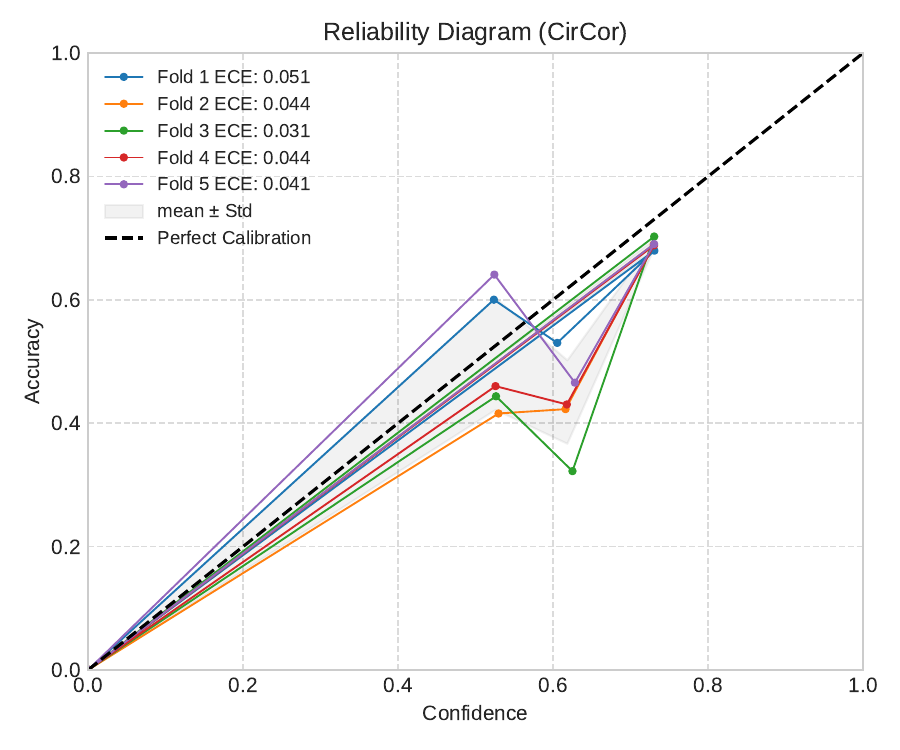}
    \caption{}
\end{subfigure}
\caption{Reliability diagrams comparing model calibration on the CinC and CirCor datasets; (a) Calibration curve for the CinC dataset and (b) calibration curve for the CirCor dataset. The x-axis represents predicted confidence (mean predicted probability per bin) and the y-axis shows observed accuracy (fraction of positive cases) within each bin. The dashed diagonal line indicates perfect calibration.}
\label{fig:cinc_circor_relib}
\end{figure}

The reliability diagrams in Figure \ref{fig:cinc_circor_relib} show that the model achieves high accuracy on both datasets while maintaining moderate, uncertainty-aware confidence. Low expected calibration error (ECE) values indicate well-calibrated predictions, aligning confidence with empirical accuracy. This conservative approach is particularly beneficial for noisy and imbalanced biomedical data like PCGs, ensuring reliable predictions without overconfidence on ambiguous signals.

\begin{figure}[H]
\centering
\begin{subfigure}{0.49\textwidth}
    \centering
    \includegraphics[width=\linewidth, clip=true, trim=50 90 80 120]{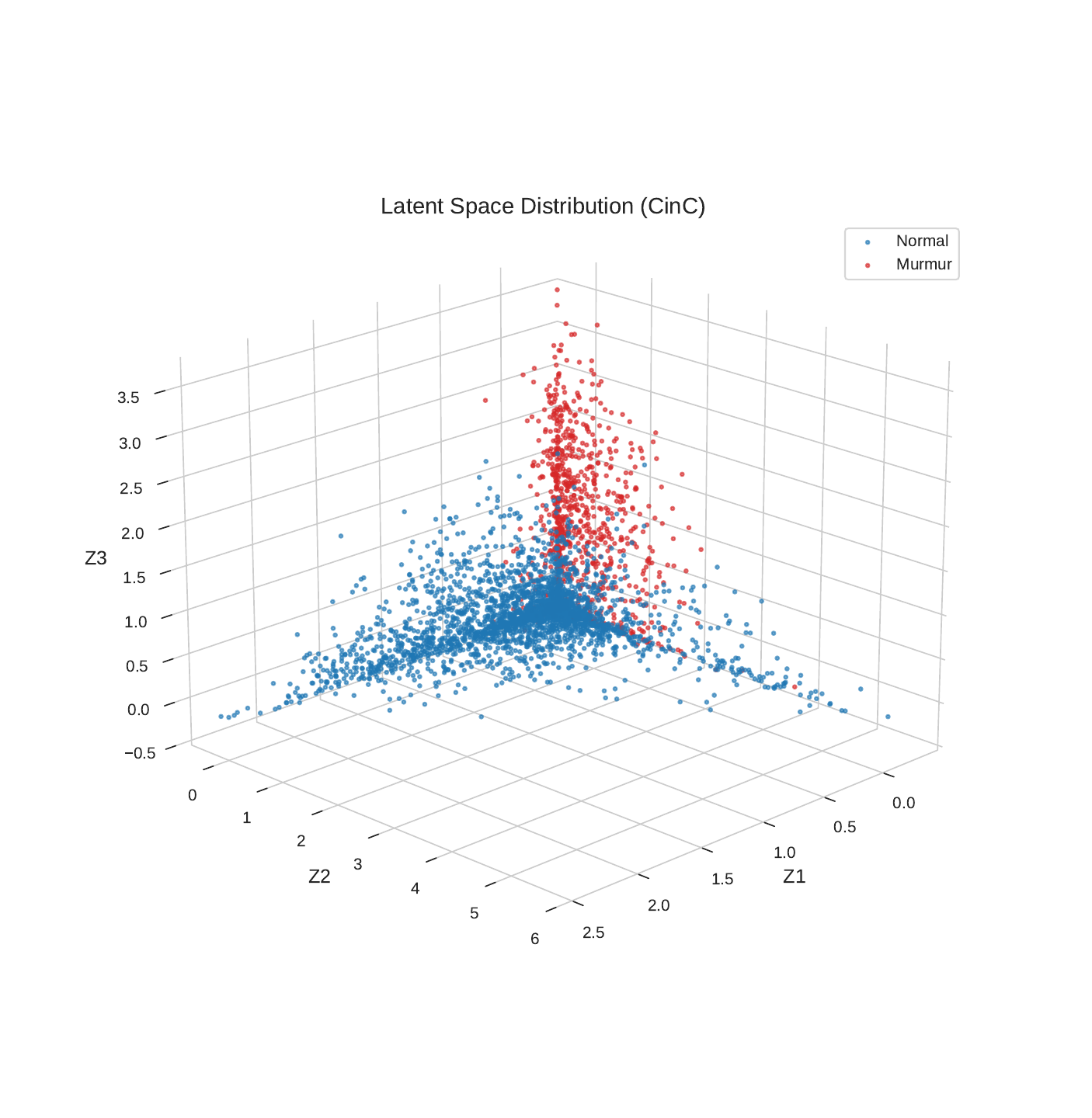}
    \caption{}
\end{subfigure}
\hfill
\begin{subfigure}{0.49\textwidth}
    \centering
    \includegraphics[width=\linewidth, clip=true, trim=50 90 80 120]{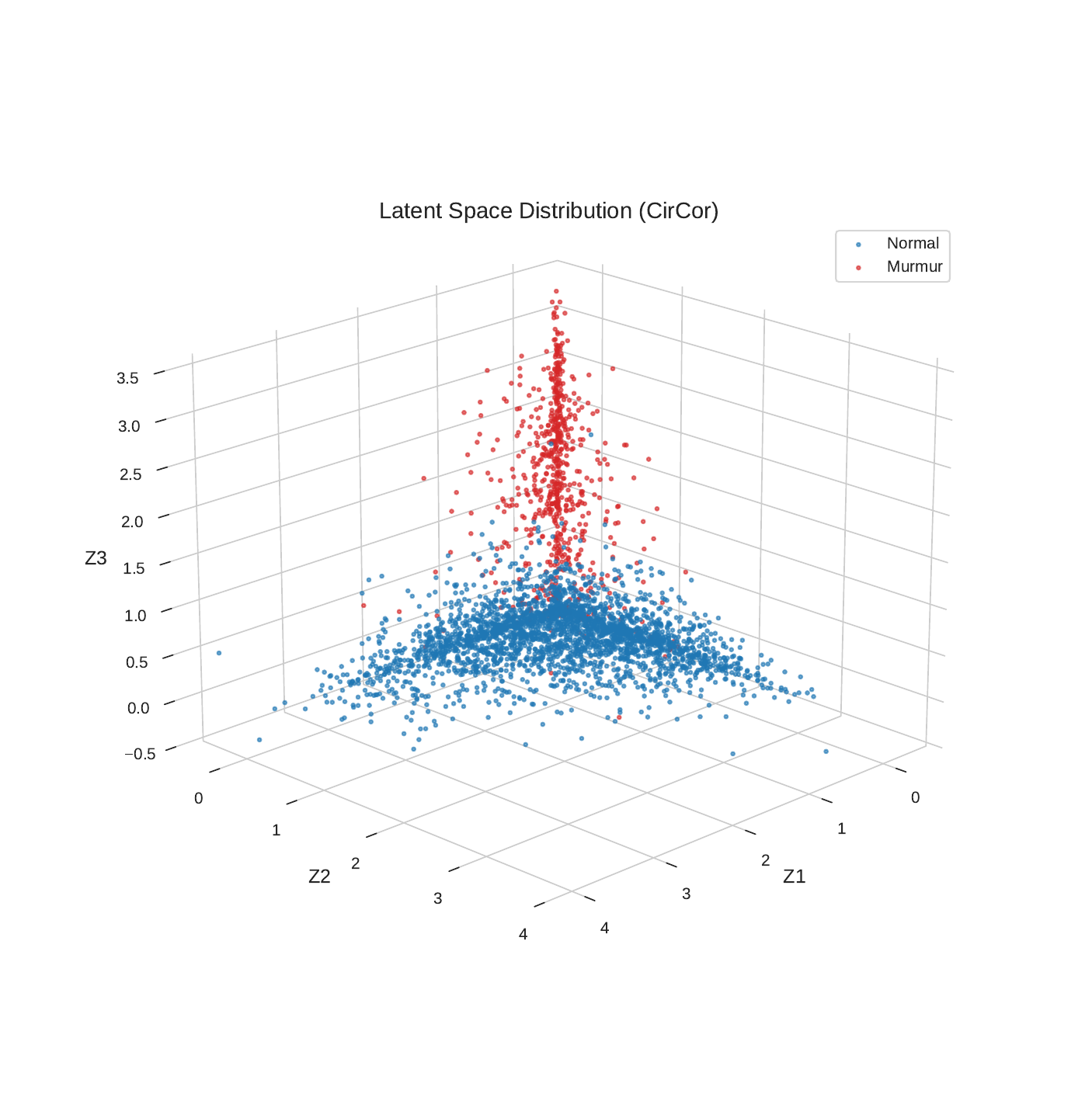}
    \caption{}
\end{subfigure}
\caption{3D latent space distribution of the model’s bottleneck features for (a) CinC and (b) CirCor datasets.}
\label{fig:cinc_circor_latent_3d}
\end{figure}

To better interpret the learned representations, the latent space of the model’s bottleneck is visualized by selecting one model from the $5$-fold cross-validation for each dataset. Figure \ref{fig:cinc_circor_latent_3d} visualizes the learned latent representations by projecting the bottleneck feature vectors onto their first three principal components ($Z_1$, $Z_2$, $Z_3$), which capture the directions of maximum variance in the embedding space. The left and right panels correspond to the CinC and CirCor datasets, respectively. The distinct clustering of Normal and Murmur samples in this reduced space indicates that QiVC-Net effectively learns discriminative, uncertainty-aware features. Note that the axes ($Z_1$, $Z_2$, $Z_3$) represent principal components derived from the bottleneck activations.

The scatter plots reveal distinct clustering patterns for normal and abnormal heart sound recordings, indicating that the model effectively captures discriminative features in its latent space. Points that lie close together represent samples with similar learned representations, while separable clusters demonstrate that the network has learned meaningful class-specific features despite noise and class imbalance.

\section{Ablation Study}

To rigorously validate the contribution of the QiRE mechanism and address concerns regarding baseline comparisons, we conducted a comprehensive ablation study within the proposed RFR architecture. This analysis isolates the effect of the structured noise strategy from other architectural components, ensuring that observed performance gains are attributable to the QiRE mechanism itself rather than incidental factors such as parameter count or regularization strength.

\subsection{Experimental Setup}

All ablation variants utilize the identical RFR backbone architecture, hyperparameters, and training protocol. Crucially, all variational variants maintain default initialization. This configuration ensures that performance differences arise solely from the sampling strategy. Consequently, the primary distinction between variants lies in the convolutional layer implementation and its associated method for weight perturbation. We evaluate five distinct variants:

\begin{enumerate}
    \item \textbf{Deterministic}: Uses standard 1D convolutional layers. This variant serves as the conventional deep learning baseline without uncertainty modeling or KL regularization.
    
    \item \textbf{Flipout} \cite{wen2018flipout}: Implements Bayesian convolution, available in TensorFlow Probability. This method employs efficient pseudo-independent weight perturbations to decorrelate gradients within a mini-batch, providing an established Bayesian CNN baseline with unbiased gradient estimation.
    
    \item \textbf{Reparameterization} \cite{kingma2013auto}: Implements variational inference which is also available in TensorFlow Probability. This variant utilizes the standard reparameterization trick for mean-field Gaussian approximations, serving as the canonical variational inference baseline.
    
    \item \textbf{Gaussian}: Utilizes our custom QiVConv layer configured to apply isotropic Gaussian noise instead of QiRE. This variant enables a direct ablation of the sampling mechanism while keeping the surrounding architecture identical to the proposed method, isolating the impact of structured rotations versus standard perturbations.
    
    \item \textbf{QiVConv}: Uses QiER method. This is the proposed method employing structured noise for geometry-preserving uncertainty quantification.
\end{enumerate}

To ensure a fair comparison, all variational variants (Flipout, Reparameterization, Gaussian, QiRE) use identical parameterization. The deterministic variant uses standard 1D convolutional layers without variational parameters or KL regularization, representing the conventional deep learning baseline. All models in the ablation study were evaluated using $5$-fold stratified cross-validation on both the CinC and CirCor datasets. We report mean $\pm$ standard deviation across folds for accuracy, sensitivity, specificity, and F1-score.

\subsection{Sampling Strategy}

Table \ref{tab:ablation_results} presents the ablation results comparing different uncertainty modeling strategies within the identical RFR architecture. The key findings are:

\begin{enumerate}
    \item \textbf{Uncertainty modeling improves predictive performance}: All variational methods (Flipout, Reparameterization, Gaussian, QiRE) outperform the deterministic baseline in accuracy, sensitivity, specificity, and F1-score, confirming that explicit modeling of parameter uncertainty improves PCG classification under noisy and imbalanced conditions.
    
    \item \textbf{Structured noise outperforms isotropic noise}: The proposed QiRE mechanism consistently achieves higher accuracy and F1-score compared to standard mean-field variational inference (Gaussian variant), suggesting that geometry-preserving rotations yield more structured and effective perturbations than independent Gaussian noise.
    
    \item \textbf{Calibration behavior differs across methods}: The inclusion of ECE reveals that improved predictive performance does not necessarily imply improved calibration. While the deterministic baseline exhibits the lowest ECE on CinC and competitive calibration on CirCor, it lacks uncertainty awareness. Among probabilistic methods, QiRE achieves competitive calibration on CirCor (matching the best ECE) while maintaining superior predictive performance. On CinC, QiRE shows slightly higher ECE compared to Gaussian and deterministic variants, indicating a trade-off between predictive accuracy and confidence calibration.
    
    \item \textbf{QiRE matches or exceeds established Bayesian baselines}: QiRE achieves comparable or superior performance to Flipout and Reparameterization baselines while maintaining identical parameterization. Notably, QiRE yields the highest accuracy and F1-score on both datasets, suggesting improved reliability for classification tasks, while maintaining competitive calibration performance.
    
    \item \textbf{Consistency across datasets}: The relative ranking of methods is consistent between CinC and CirCor in terms of predictive performance, and QiRE maintains stable calibration behavior across datasets, indicating that the benefits of structured uncertainty modeling generalize across different PCG acquisition conditions and patient populations.
\end{enumerate}

\begin{table}[H]
\centering
\caption{Ablation study of sampling strategies within the RFR architecture. Results are mean $\pm$ std across 5-fold stratified cross-validation.}
\label{tab:ablation_results}
\begin{adjustbox}{max width=\textwidth}
\begin{tabular}{l M{2.2cm} M{2.1cm} M{2.2cm} M{2.2cm} M{2.0cm}  M{1.85cm}}
\hline
\textbf{Dataset} & \textbf{Variant} & \textbf{Accuracy (\%)} & \textbf{Sensitivity (\%)} & \textbf{Specificity (\%)} & \textbf{F1-Score (\%)}& \textbf{ECE $\downarrow$} \\
\hline
\multirow{5}{*}{\textbf{CinC}} 
& Deterministic      & 95.99 $\pm$ 1.09 & 92.34 $\pm$ 1.87 & 97.21 $\pm$ 1.56 & 91.53 $\pm$ 1.73 & \textbf{0.054 $\pm$ 0.02} \\
& Flipout            & 96.73 $\pm$ 0.98 & 93.63 $\pm$ 1.62 & 97.74 $\pm$ 1.12 & 92.99 $\pm$ 1.49 & 0.083 $\pm$ 0.01 \\
& Reparameterization & 95.46 $\pm$ 0.91 & 92.08 $\pm$ 1.68 & 96.58 $\pm$ 1.24 & 90.27 $\pm$ 1.67 & 0.102 $\pm$ 0.02 \\
& Gaussian           & 95.03 $\pm$ 1.31 & 90.91 $\pm$ 1.94 & 96.36 $\pm$ 1.72 & 89.36 $\pm$ 1.84 & 0.071 $\pm$ 0.03 \\
& \textbf{QiRE}   & \textbf{97.84 $\pm$ 0.91} & \textbf{96.89 $\pm$ 1.91} & \textbf{98.14 $\pm$ 1.34} & \textbf{95.60 $\pm$ 1.67} & 0.094 $\pm$ 0.01 \\
\hline
\multirow{5}{*}{\textbf{CirCor}} 
& Deterministic      & 96.30 $\pm$ 1.43 & 93.28 $\pm$ 2.06 & 97.31 $\pm$ 0.79 & 92.05 $\pm$ 2.67 & \textbf{0.042 $\pm$ 0.05} \\
& Flipout            & 96.94 $\pm$ 0.99 & \textbf{94.50 $\pm$ 2.37} & 97.75 $\pm$ 0.82 & 93.49 $\pm$ 1.94 & 0.050 $\pm$ 0.03 \\
& Reparameterization & 95.67 $\pm$ 1.52 & 91.97 $\pm$ 3.19 & 96.86 $\pm$ 0.68 & 90.54 $\pm$ 2.70 & 0.076 $\pm$ 0.01 \\
& Gaussian           & 94.95 $\pm$ 1.16 & 90.41 $\pm$ 2.53 & 96.45 $\pm$ 0.72 & 89.19 $\pm$ 2.55 & 0.168 $\pm$ 0.11 \\
& \textbf{QiRE}   & \textbf{97.89 $\pm$ 1.21} & 92.34 $\pm$ 3.05 & \textbf{99.27 $\pm$ 0.64} & \textbf{95.55 $\pm$ 2.88} & \textbf{0.042 $\pm$ 0.01} \\
\hline
\end{tabular}
\end{adjustbox}
\end{table}

\subsection{Subspace Dimension Sensitivity Analysis}

The QiRE mechanism introduces a hyperparameter $k$, the dimension of the random subspace used for unitary rotations. To assess the robustness of our method to this choice and provide guidance for practitioners, we conducted a sensitivity analysis over $k \in \{1, 3, 5, 7, 9, 11, 15, 30\}$. All other hyperparameters and the training protocol were held constant.

For computational efficiency, this sensitivity analysis was conducted using a single fixed stratified train/test split ($80\%$/$20\%$) rather than $5$-fold cross-validation. Training employed an early stopping mechanism to prevent from overfitting by monitoring the validation F1-score, with a patience of $30$ epochs. Training was terminated if no improvement over the best observed score occurred within this window. Table \ref{tab:sensitivity_k} shows the impact of $k$ on classification accuracy and the number of training epochs (determined by early stopping). The results indicate:

\begin{enumerate}
    \item \textbf{Stable performance for moderate $k$}: Accuracy remains relatively stable for $k \in [3, 9]$, suggesting that the method is robust to moderate variations in subspace dimensionality.
    
    \item \textbf{Diminishing returns for large $k$}: Performance degrades for $k \geq 11$, with accuracy dropping notably at $k=30$. This is likely due to increased computational noise in high-dimensional rotations and potential over-regularization, as reflected by the higher epoch counts required for convergence.
    
    \item \textbf{Minimal $k$ is insufficient}: The extreme case $k=1$ yields lower accuracy compared to $k=5$, confirming that a minimal subspace cannot capture the structured correlations needed for effective uncertainty modeling.
    
    \item \textbf{Default choice justified}: Our default setting $k=5$ achieves the highest accuracy on both datasets with moderate epoch counts, suggesting the empirical choice reported in the main experiments.
\end{enumerate}

\begin{table}[H]
\centering
\caption{Sensitivity analysis of subspace dimension $k$ in QiRE.}
\label{tab:sensitivity_k}
\begin{adjustbox}{max width=\textwidth}
\begin{tabular}{M{1.0cm} M{2.2cm} M{2.2cm} M{2.2cm} M{2.2cm}}
\hline
\multirow{2}{*}{\textbf{$k$}} & \multicolumn{2}{c}{\textbf{CinC}} & \multicolumn{2}{c}{\textbf{CirCor}} \\
\cline{2-5}
& \textbf{Acc (\%)} & \textbf{Epochs} & \textbf{Acc (\%)} & \textbf{Epochs} \\
\hline
1 & 95.73 & 78 & 95.92 & 90 \\
3 & 95.08 & 77 & 96.18 & 92 \\
5 & \textbf{96.51} & 79 & \textbf{97.35} & 91 \\
7 & 96.36 & 65 & 96.97 & 105 \\
9 & 94.49 & 57 & 95.24 & 69 \\
11 & 93.85 & 134 & 93.99 & 65 \\
15 & 93.80 & 115 & 94.35 & 83 \\
30 & 91.77 & 96 & 92.55 & 111 \\
\hline
\end{tabular}
\end{adjustbox}
\end{table}

To quantify the computational overhead associated with the subspace dimension $k$, we conducted a controlled timing analysis using $100$ random samples from a representative subset of the training data, averaging results over three independent runs of 10 epochs each. Our empirical results demonstrate a proportional scaling behavior. Specifically, increasing $k$ from $1$ to $30$ resulted in a training time increase of approximately $2.4$ times. This finding underscores the necessity of selecting a moderate $k$ value to optimize the critical trade-off between maximizing predictive accuracy and maintaining computational efficiency, particularly for resource-constrained deployments.

\subsection{Discussion}

The ablation results confirm that the QiRE mechanism provides measurable benefits beyond standard variational inference and established Bayesian CNN baselines, demonstrating that the observed performance gains are attributable to the structured rotation-based sampling strategy rather than generic variational regularization. 

Importantly, the inclusion of calibration metrics (ECE) provides additional insight into the behavior of different uncertainty modeling approaches. While deterministic models can exhibit low ECE due to overconfident predictions aligned with dominant classes, they lack the ability to represent epistemic uncertainty. In contrast, variational methods introduce stochasticity that may slightly increase ECE but provide more informative predictive distributions. Within this context, QiRE achieves a favorable balance: it delivers superior accuracy and F1-score while maintaining competitive calibration, particularly on the CirCor dataset where it achieves the lowest ECE among all probabilistic methods.

While Flipout yields marginally higher sensitivity on the CirCor dataset, QiRE demonstrates improved overall accuracy, F1-score, and specificity, indicating a more balanced performance profile suitable for clinical decision support. Additionally, the sensitivity analysis shows that QiRE is robust to the choice of subspace dimension $k$, with stable performance across a reasonable range ($k \in [1, 9]$), reducing the burden of hyperparameter tuning and supporting practical deployment.

Overall, these results suggest that structured, geometry-preserving perturbations can improve predictive performance while maintaining reliable uncertainty estimates, offering a practical advantage over conventional mean-field variational approaches.

\section{Limitations and Future Work}

While QiVC-Net demonstrates robust performance in uncertainty-aware PCG classification, certain aspects offer opportunities for further refinement. First, although the QiRE mechanism is mathematically grounded in unitary evolution (as detailed in Appendix A), its current implementation relies on a fixed subspace dimension $k$. While our sensitivity analysis identifies an optimal range ($k \in [3, 9]$), future work could explore adaptive mechanisms to dynamically adjust $k$ based on layer depth or data complexity, potentially enhancing efficiency without manual tuning. Second, although the decoherence step is optional while provides effective regularization, it introduces an additional hyperparameter ($p$) that requires dataset-specific calibration. Investigating self-tuning strategies for this parameter remains a promising direction. 

Finally, although QiVC-Net maintains a parameter count comparable to BNNs, the stochastic sampling process incurs a modest computational overhead during training compared to standard forward passes. Additionally, a limitation of the current study is that the framework has not yet been evaluated against transformer-based models.

\section{Conclusion}

In this study, QiVC-Net, a variational convolutional network drawing a mathematical analogy from quantum transformations, was introduced for the classification of PCG signals. The model demonstrated consistently high performance across both the CinC and CirCor datasets, achieving average accuracies close to $98\%$, strong specificity, and competitive sensitivity, while effectively handling class imbalance. Robustness analysis under varying noise conditions indicated that discriminative capability and confidence calibration were maintained even with degraded input signals. Reliability diagrams confirmed that the model produced well-calibrated predictions, reflecting a conservative and uncertainty-aware behavior. Using a composite loss function with dynamic weighting, along with early stopping and stratified cross-validation, the model achieved stable and generalizable performance across folds. Comparative analysis with other methods, such as those in previous studies \cite{soares2020autonomous, ghosh2022automated, patwa2025heart}, shows that QiVC-Net outperforms existing approaches while maintaining balanced sensitivity and specificity. Overall, these results demonstrate that variational, uncertainty-aware convolutional architectures are effective for robust classification in challenging, noisy, and corrupted biomedical datasets, including PCG signals. The model maintained reliable performance even under conditions of data imbalance and signal degradation.

The QiVC framework offers several notable advantages. By introducing a geometry-preserving uncertainty modeling approach through the QiRE mechanism, the framework enables the model to capture structured variability in the convolutional weights while maintaining stable training dynamics. The design remains parameter-efficient, as it does not add additional learnable weights beyond the mean and standard deviation, making the approach suitable for lightweight deployment. Moreover, QiVC leads to improved prediction calibration and helps prevent over-confidence, particularly in noisy or ambiguous cases, an important factor in clinical and safety-critical applications. While QiVC demonstrates clear benefits in uncertainty modeling and calibration, some aspects remain open for refinement. The quantum-inspired rotational ensemble in QiRE is currently motivated conceptually rather than through a complete theoretical formalization, and the selection of its subspace dimension may require empirical adjustment across datasets. Moreover, as this is the first instance of applying such a structured probabilistic convolution, broader ablation studies will help further characterize its behavior and contribution relative to conventional stochastic or variational layers. These limitations do not fundamentally hinder the method but rather outline natural directions for continued development and validation.

\section*{Declaration of Competing Interest}

The authors declare that they have no known competing financial interests or personal relationships that could have appeared to influence the work reported in this paper.

\section*{Authorship Contribution Statement}

\textbf{Amin Golnari:} Conceptualization, Methodology, Software, Investigation, Visualization, Writing – Original Draft, Writing – Review \& Editing. \textbf{Jamileh Yousefi:} Supervision, Validation, Writing – Review \& Editing. \textbf{Reza Moheimani:} Investigation, Writing – Review \& Editing. \textbf{Saeid Sanei:} Supervision, Validation, Formal Analysis, Writing – Review \& Editing.

\bibliographystyle{unsrturl}  
\bibliography{references}


\section*{Appendix A: Quantum-Inspired Foundations of the QiVC Transformation}

The QiVC perturbation mechanism is inspired by two foundational mathematical structures in quantum mechanics:  
(1) unitary evolution, which preserves the norm (total probability) of a quantum state, and  
(2) projective decomposition, which expresses a state relative to a chosen measurement basis.  
QiVC does not simulate a quantum system; rather, it imports these structures to construct stable and directionally controlled stochastic perturbations in neural representations.  
Formal treatments of these concepts appear in \textbf{Sections 1.4, 1.5, and 2.2.1 of \cite{sakurai2020modern}}.

\subsection*{A.1 Norm Preservation and Unitary Analogy}

In quantum mechanics, a pure state is denoted by the ket vector $|\psi\rangle$, which satisfies the normalization condition:
\begin{equation}\label{eq:app_norm}
\| |\psi\rangle \| = 1.
\end{equation}
Its time evolution is governed by a unitary operator $U$, satisfying the condition $U^\dagger U = I$ (where $U^\dagger$ is the conjugate transpose and $I$ is the identity matrix). The evolution preserves the state norm:
\begin{equation}\label{eq:app_unitary}
\|U|\psi\rangle\| = \| |\psi\rangle \|.
\end{equation}
This ensures that transformations change the orientation of the state vector but not its magnitude (see \textbf{Sec. 1.5.1} of \cite{sakurai2020modern}).

\paragraph{Correspondence in QiVC.}  
QiVC enforces that perturbations modify feature \emph{direction}, not \emph{scale}, to maintain stable representational magnitude.
A perturbation is initialized by sampling a noise vector $\epsilon_0$ from a standard normal distribution $\mathcal{N}(0, I_N)$, where $N$ is the total number of kernel parameters. This vector is then normalized to unit length:
\begin{equation}\label{eq:app_norm_qivc}
\epsilon = \frac{\epsilon_0}{\|\epsilon_0\|}.
\end{equation}

\subsection*{A.2 Projective Decomposition (Measurement Analogy)}

A $k$-dimensional subspace (where $k \ll N$) is constructed using a random basis matrix $Q \in \mathbb{R}^{N \times k}$. $Q$ is obtained via the thin QR decomposition of a random Gaussian matrix $G \in \mathbb{R}^{N \times k}$:
\begin{equation}\label{eq:app_qr}
G = QR, \qquad \text{such that } Q^\top Q = I_k,
\end{equation}
where $I_k$ is the $k \times k$ identity matrix.
We define two projection operators based on $Q$: the parallel projector $P_{\parallel} = Q Q^\top$ and the orthogonal projector $P_{\perp} = I - P_{\parallel}$.
The normalized noise vector $\epsilon$ is then decomposed into two orthogonal components:
\begin{equation}\label{eq:app_decomp}
\epsilon_{\parallel} = P_{\parallel}\epsilon, \qquad 
\epsilon_{\perp} = P_{\perp}\epsilon,
\end{equation}
satisfying the orthogonality condition $\epsilon_{\parallel}^\top \epsilon_{\perp} = 0$.

\paragraph{Correspondence in QiVC.}  
This mirrors the decomposition of a quantum state into measured and unmeasured components under projective measurement  
(see \textbf{Sec. 1.4, Projection Postulate}, \cite{sakurai2020modern}).

\subsection*{A.3 Rotation Within the Subspace (Change of Basis Analogy)}

A rotation is applied exclusively to the parallel component $\epsilon_{\parallel}$ using a special orthogonal matrix $U \in \mathrm{SO}(k)$, which is sampled from the Haar measure over the group of $k \times k$ rotations. The rotated component is computed as:
\begin{equation}\label{eq:app_rot}
\epsilon_{\mathrm{rot}} = Q U Q^\top \epsilon.
\end{equation}
The final perturbation vector $\epsilon_{\mathrm{final}}$ is reconstructed by recombining the unchanged orthogonal component $\epsilon_{\perp}$ with the rotated parallel component:
\begin{equation}\label{eq:app_final}
\epsilon_{\mathrm{final}} = \epsilon_{\perp} + \epsilon_{\mathrm{rot}}.
\end{equation}

\paragraph{Correspondence in QiVC.}  
This corresponds to applying a unitary basis transformation within a subspace, which leaves magnitude unchanged but redistributes uncertainty  
(see \textbf{Sec. 1.5, Basis Transformations}, \cite{sakurai2020modern}).

\subsection*{A.4 Verification of Norm Preservation (Proof)}

Using the orthonormality property $Q^\top Q = I_k$ and the unitary property $U^\top U = I_k$, the squared norm of the rotated component is:
\begin{align}\label{eq:app_proof1}
\|\epsilon_{\mathrm{rot}}\|^2 &= (Q U Q^\top \epsilon)^\top (Q U Q^\top \epsilon) \\
&= \epsilon^\top Q U^\top (Q^\top Q) U Q^\top \epsilon \\
&= \epsilon^\top Q (U^\top U) Q^\top \epsilon \\
&= \epsilon^\top Q Q^\top \epsilon \\
&= \|\epsilon_{\parallel}\|^2.
\end{align}
Since $\epsilon_{\parallel}$ and $\epsilon_{\perp}$ are orthogonal, the Pythagorean theorem applies to the final vector:
\begin{equation}\label{eq:app_proof2}
\|\epsilon_{\mathrm{final}}\|^2 = \|\epsilon_{\parallel}\|^2 + \|\epsilon_{\perp}\|^2 = \|\epsilon\|^2 = 1.
\end{equation}
Thus, the norm is strictly preserved ($\|\epsilon_{\mathrm{final}}\| = 1$).

\subsection*{A.5 Operator Form and Unitary Analogue}

We define the full QiVC transformation operator $R_{\mathrm{QiVC}}$ acting on the noise vector as:
\begin{equation}\label{eq:app_operator}
R_{\mathrm{QiVC}} = P_{\perp} + Q U Q^\top.
\end{equation}
Using the properties of projectors ($P_{\perp}^2 = P_{\perp}$, $P_{\perp} P_{\parallel} = 0$) and the unitarity of $U$, it can be shown that:
\begin{equation}\label{eq:app_orthogonal}
R_{\mathrm{QiVC}}^\top R_{\mathrm{QiVC}} = I.
\end{equation}

\paragraph{Correspondence in QiVC.}  
This demonstrates that $R_{\mathrm{QiVC}}$ is the real-valued analogue of a \emph{unitary operator}, preserving quantum probability amplitudes  
(see \textbf{Sec. 2.2.1}, \cite{sakurai2020modern}).

\subsection*{A.6 Interpretation: Quantum-Inspired Uncertainty Redistribution}

Because the rotation matrix $U$ and the projection operator $P_{\parallel}$ do not necessarily commute in broader contexts, QiVC reorients the uncertainty distribution without altering its total magnitude.  
This parallels the uncertainty trade-offs arising from non-commuting observables in quantum mechanics  
(see \textbf{Sec. 1.4.5}, \cite{sakurai2020modern}).

\end{document}